\documentclass[12pt,draftcls,onecolumn]{IEEEtran}
\usepackage{cite}
\usepackage{amsthm}
\usepackage{amsmath,amssymb,amsfonts}
\usepackage{graphicx,float}
\usepackage{hyperref}
\usepackage{subfig}
\usepackage{tcolorbox}
\usepackage{algorithm}
\usepackage{algpseudocode}
\hypersetup{hidelinks=true}
\usepackage{textcomp}
\def\BibTeX{{\rm B\kern-.05em{\sc i\kern-.025em b}\kern-.08em
    T\kern-.1667em\lower.7ex\hbox{E}\kern-.125emX}}
\newtheorem{theorem}{Theorem}[section]
\newtheorem{proposition}[theorem]{Proposition}
\newtheorem{lemma}[theorem]{Lemma}
\newtheorem{corollary}[theorem]{Corollary}
\newtheorem{definition}[theorem]{Definition}
\newtheorem{assumption}[theorem]{Assumption}
\newtheorem{remark}[theorem]{Remark}
\begin{document}
\title{Mechanism Design for Heterogenous Differentially Private Data Acquisition for Logistic Regression}
\author{Ameya Anjarlekar, Rasoul Etesami, and R. Srikant
\vspace{-0.2cm}
\thanks{Ameya Anjarlekar and R. Srikant are with Department of Electrical and Computer Engineering and Coordinated Science Lab, University of Illinois at Urbana-Champaign, USA
        {\tt\small ameyasa2@illinois.edu; rsrikant@illinois.edu}}%
\thanks{Rasoul Etesami is with the Departments of Industrial and Systems Engineering and  Electrical and Computer Engineering and the Coordinated Science Lab, University of Illinois Urbana-Champaign, USA
        {\tt\small etesami1@illinois.edu}}%
        \thanks{The work done in this paper was supported by NSF Grants CCF 22-07547, CCF 1934986, CNS 21-06801, CAREER Award EPCN-1944403, AFOSR Grant FA9550-23-1-0107, and ONR Grant N00014-19-12566.}
}
\maketitle
\begin{abstract}
We address the challenge of solving machine learning tasks using data from privacy-sensitive sellers. Since the data is private, we design a data market that incentivizes sellers to provide their data in exchange for payments. Therefore our objective is to design a mechanism that optimizes a weighted combination of test loss, seller privacy, and payment, striking a balance between building a good privacy-preserving ML model and minimizing payments to the sellers. To achieve this, we first propose an approach to solve logistic regression with known heterogeneous differential privacy guarantees. Building on these results and leveraging standard mechanism design theory, we develop a two-step optimization framework. We further extend this approach to an online algorithm that handles the sequential arrival of sellers.
\end{abstract}


\section{Introduction}

Machine learning (ML) applications have experienced significant growth in recent years with applications in field ranging from large language models \cite{Brendan} to LQ control problems \cite{Yazdani}. However, the use of sensitive data such as health data or financial data has raised privacy concerns \cite{kushmaro_2021}. To solve this problem, substantial efforts have been dedicated to ensuring the privacy of training data, with the prevalent adoption of differential privacy. While existing literature presents various algorithms to guarantee differential privacy such as \cite{Degue} proposing a differentially private Kalman filtering approach, a lingering question persists: determining the optimal degree of differential privacy. For instance, opting for a higher level of differential privacy may compromise the performance of the machine learning model, yet it significantly enhances privacy protection for the data provider. Therefore, along with considering the model performance through metrics such as misclassification loss, we need to also consider the privacy loss of the data providers (also referred by sellers). In this paper, we delve into addressing this nuanced tradeoff by formulating a mechanism that balances competing objectives: achieving a high-quality ML model while minimizing the privacy loss experienced by data providers. 

To further motivate our problem, it is crucial to recognize that providing differential privacy guarantees to data providers addresses their privacy concerns to some extent, but does not automatically ensure their willingness to share their data. Data providers may also want to monetize their data since they know that their data is valuable to a machine learning model designer \cite{kushmaro_2021}. Examples where users's data is needed to build a machine model include autonomous systems and healthcare applications. Data from human drivers are often used to train machine learning models designed to provide autonomous driving features in cars\cite{waymo}. However, many drivers may be unwilling to share their driving data without sufficient compensation due to privacy concerns. As a result, self-driving car companies may need to offer financial incentives in exchange for access to this data. Moreover, each patient may have a different cost for the same loss of privacy (which we term privacy sensitivity) and therefore, may have to be compensated differently. Another example is one where a hospital aims to use patients' health vitals to predict heart disease. Even with robust privacy guarantee, patients may still hesitate to share such sensitive data unless they are adequately compensated.  To address these issues, there is a growing interest in encouraging data sharing through two strategies: (i) introducing noise to ML model's weights to enhance dataset anonymity, and (ii) providing compensation to data sellers to offset potential privacy risks \cite{posner}. The amount of compensation that is provided to patients would, of course, depend on their privacy sensitivity and their privacy loss, which can be measured using differential privacy. To operationalize this concept and thus accurately represent the tradeoff between model performance and privacy loss, we propose a robust mathematical framework for designing a data market. This market facilitates data acquisition from privacy-conscious sellers, utilizing (a) mechanism design to incentivize sellers to truthfully disclose their privacy sensitivity (we consider that sellers can lie about their privacy sensitivities) and (b) statistical learning theory to strike a balance between payments and model accuracy. The market involves a buyer seeking data from privacy-sensitive sellers, with the aim of building a high-quality ML model while minimizing overall payments to sellers. Conversely, individual sellers seek fair compensation for potential privacy compromises. Additionally, we note that in reality, sellers might not be comfortable revealing their data without first receiving a payment. To achieve this, we need to impose additional constraints on our model in which payments and privacy guarantees need to be computed without the knowledge of the sellers' data. Incorporating all these requirements, in Section \ref{section: mech design} and Section \ref{sec: algo} we construct a model for the data market wherein we capture the tradeoff between designing a good ML model consistent with the privacy guarantees while ensuring small privacy loss to sellers.

Additionally, we consider another scenario where sellers may approach the platform sequentially. Since sellers can be impatient, the buyer must decide on payments and differential privacy guarantees immediately after the seller discloses their privacy sensitivity. Unlike traditional online decision-making problems, where there is typically weak correlation between current and past decisions, our problem is characterized by strong interdependence—decisions made at step $t$ have a significant impact on future decisions. To address this, in Section \ref{section: online setting}, we propose an algorithm tailored to solving the online variant of the mechanism design problem.\footnote{A shorter version of this work dealing with only the online version of our results given in Section \ref{section: online setting} has been accepted for publication at the IEEE Conference on Decision and Control (CDC) 2024.}

We are motivated by the work in \cite{Asu}, which considers the mean estimation of a scalar random variable using data from privacy-sensitive sellers. Our objective here is to design a mechanism for the more challenging and practically useful logistic regression problem with vector-valued data. Now, to consider the tradeoff between payments and model accuracy, it is imperative to mathematically represent the buyer's objective, i.e., model accuracy. In contrast to \cite{Asu}, where the buyer's objective simplifies to the variance of a mean estimator, which they assume to be known, our scenario considers the buyer's objective to be the expected misclassification error of a logistic regression model in which the statistics of the dataset are unknown. To address this issue, we propose using Rademacher complexity to model the buyer's objective. Furthermore, most prior work on differentially private logistic regression, such as \cite{kamalika} and \cite{Ding}, consider homogeneous differential privacy, in which every individual has the same privacy guarantees. However, our approach acknowledges the practical reality that sellers might have different degrees of willingness (privacy sensitivity) to share their data. Therefore, we consider that each data point has to be protected differently through heterogeneous differential privacy, leading to different utility of each data point in contributing to the ML model.

To summarize, our goal is to design a mechanism for the buyer to optimize an objective that trades off between classification loss and payments to sellers while also taking into account the differential privacy requirements of the sellers.

\subsection{Contributions and Organization}
\begin{itemize}
\item In Section \ref{sec: regression}, we provide an approach to model the misclassification loss with heterogenous differential privacy guarantees for logistic regression. We further highlight that unlike the case of the same differential privacy for all users as in \cite{kamalika} where the addition of a single noise term to the objective is sufficient to provide privacy, heterogeneous privacy requires additional nuances such as considering the generalization bounds of the regression model which we derive using ideas from statistical learning theory.

\item Building on this result, we solve the mechanism design problem (Section \ref{section: mech design}). For this problem, we provide a payment identity that determines payments as a function of the privacy guarantees. This, along with results from Section \ref{sec: regression} is used to design an objective for the mechanism design problem. Subsequently, we propose an algorithm to solve the mechanism design problem.


\item We also conduct an asymptotic analysis in Section \ref{sec: algo} by considering a large number of sellers to highlight the privacy-utility tradeoff of our algorithm which is further used to understand how much it will cost a buyer to obtain sufficient data to ensure a certain misclassification loss in the ML model that results from the mechanism. The interesting insight here is that, because the buyer can selectively choose sellers to acquire data from, the budget required for a given bound on the misclassification loss is bounded. 

\item We further extend the algorithm proposed in Section \ref{sec: algo} to solve an online setting wherein sellers arrive sequentially to the buyer/platform and report their privacy sensitivity. This extension addresses challenges such as need for immediate decision-making on payments and differential privacy guarantees, especially when sellers expect upfront payments. We tackle the history-dependence issue of this problem by developing an approximate algorithm.

\item Finally, we demonstrate the application of our proposed mechanism on the Wisconsin breast cancer data set \cite{data}. We observe fast convergence, indicating the usefulness of the change of variables. Additionally, we evaluate the performance of both online and offline algorithms. 
\end{itemize}

The paper is structured as follows: In Section \ref{sec: regression}, we present a novel approach to solving logistic regression under heterogeneous differential privacy guarantees, a key result for addressing the mechanism design problem. Section \ref{section: mech design} introduces the mechanism design problem in detail, and Section \ref{sec: algo} proposes an asymptotically optimal algorithm to solve it. Section \ref{section: online setting} extends the problem to an online setting. Finally, Section \ref{sec: num} validates the theoretical results by applying our method to a medical dataset. All the proofs are deferred to the appendix. The appendix also contains additional experiments to further illustrate and provide intuition for the solution. We also discuss computational techniques to solve the mechanism design objective in Section \ref{sec: data dependent pay}.

\subsection{Related Work}

\textbf{Differentially Private ML Algorithms:} While literature on creating differentially private data markets is relatively sparse, there is a vast literature on incorporating differential privacy in statistical modeling and learning tasks. For example, \cite{Cummings} builds a linear estimator using data points so that there is a discrete set of privacy levels for each data point. \cite{Nissim,Ghosh3,Nissim2,Roth} use differential privacy to quantify loss that sellers incur when sharing their data. Works such as \cite{Alaggan,Nissim2,Wang,Liao,Ding} also consider problems concerned with ensuring differential privacy. Some works consider a different definition of privacy. \cite{Roth2}, \cite{Chen}, and \cite{Chen_2019} use a menu of probability-price pairs to tune privacy loss and payments to sellers. \cite{Duan,ZhangZuo} also consider problems associated with privacy-utility tradeoff. \cite{Perote,DEKEL2010759,MEIR2012123,Ghosh3,Yang} consider that sellers can submit false data. In the context of differentially private ML algorithms, a portion of our work can be viewed as contributing to differentially private logistic regression with heterogeneous sellers.

\textbf{Mechanism Design and Online Algorithms:} Mechanism design has a long history, originally in economics and more recently in algorithmic game theory. Recent work such as \cite{Rachel} considers auctions in which buyers bid multiple times. Similarly, \cite{Gallien} considers an online setting that involves a monopolist selling multiple items to buyers arriving over time. \cite{Chen} provides a mechanism that considers minimizing the worst-case error of an unbiased estimator while ensuring that the cost of buying data from sellers is small. However, in their paper, the cost is chosen from a discrete set of values. Utilizing machine learning techniques, \cite{Hartline} reduces the mechanism design problem to a standard algorithmic design problem. \cite{Montazeri} proposes a profit maximization mechanism for task allocation problems while ensuring truthful participation. Other papers, such as \cite{Ghosh2,Liu2,Nicole,Navabi}, also consider mechanism design for different objectives and problems of interest. However, none of these works incorporates ML algorithms or differential privacy in their analysis. Our work is more closely related to \cite{Asu}, where the authors develop a mechanism to estimate the mean of a scalar random variable by collecting data from privacy-sensitive sellers. However, unlike \cite{Asu}, where they assume some statistical knowledge of the quantity to be estimated, our challenge is to design a mechanism without such knowledge. This leads to interesting problems in both deriving a bound for the misclassification loss and solving a non-convex optimization problem to implement the mechanism.


\section{Logistic Regression while Ensuring Heterogeneous Differential Privacy}
\label{sec: regression}
A major challenge in designing the optimal mechanism is to represent the misclassification error for the problem of logistic regression with heterogenous differential privacy constraints. In this section, we first formally define differential privacy and then introduce the problem of representing the misclassification error.

\subsection{Differential Privacy}
To build the necessary foundation, we define the notion of privacy loss that we adopt in this paper. We assume that sellers trust the platform to add necessary noise to the model weights to keep their data private. This is called central differential privacy. The first definition of differential privacy was introduced by \cite{Dwork1}, which considered homogeneous differential privacy (i.e., same privacy guarantee to all data providers). In our paper, we consider a slight extension wherein the users are provided different privacy guarantees, i.e., \emph{heterogeneous} differential privacy. More formally, it is defined as follows.

\begin{definition}
Let $\boldsymbol{\epsilon} = (\epsilon_i)_{i=1}^{m} \in \mathbb{R}_{+}^{m}$. Also, consider $S$, $S'$ to be two datasets that differ in $i^{th}$ component with $|S| = |S'| = m$. Let  $\mathbb{A}$ be an algorithm that takes a dataset as input and provides a vector in $\mathbb{R}^n$ as output. We say that $\mathbb{A}$ provides $\boldsymbol{\epsilon}$-centrally differential privacy, if for any set $V \subset  \mathbb{R}^n,$
\begin{equation}
    e^{-\epsilon_i} \leq \frac{\mathbb{P}[\mathbb{A}(S) \in V]}{\mathbb{P}[\mathbb{A}(S') \in V]} \leq e^{\epsilon_i} \quad \forall i \in \{1,2,\ldots,m\}.
\end{equation}
\end{definition}

\medskip
This definition states that if the value of $\epsilon_i$ is small, then it is difficult to distinguish between the outputs of the algorithm when the data of seller $i$ is changed. Note that a smaller value of $\epsilon_i$ means a higher privacy guarantee for the seller.

\subsection{Representing the Misclassification Error}

Before introducing our mechanism design problem, we will first focus on a simpler problem, i.e., logistic regression while providing heterogeneous differential privacy requirements. The results of this subsection will later be used when we consider the mechanism. We consider the following problem:
\begin{itemize}
    \item We have a set of $m$ users, with users having iid data points $\boldsymbol{z}^i = (\boldsymbol{x}^i,y^i),$ with $n$-dimensional input $\boldsymbol{x}$ and output $y$ sampled from a joint distribution $\mathcal{D}(\boldsymbol{x}, y)$ where we assume $\|\boldsymbol{x}^i\| \leq 1 \ \forall i$. We let $D=\{(\boldsymbol{x}^1,y^1),\ldots,(\boldsymbol{x}^m,y^m)\}$ denote the data set.
    \item Each user $i$ demands that $\epsilon_i$ differential privacy must be ensured for their data.
    \item The platform aims to design the best linear estimator $\boldsymbol{w}$ by minimizing the misclassification loss $\mathbb{E}[\mathbb{I}_{\{sign(\boldsymbol{w}^T \boldsymbol{x}) \neq y\}}]$ such that $\|\boldsymbol{w}\|\!\leq\! \beta$ while ensuring differential privacy $\boldsymbol{\epsilon}$.\footnote{$\|\boldsymbol{w}\|$ is the $l_2$ norm of $\boldsymbol{w}$, and $\mathbb{I}_{\{\cdot\}}$ denotes the indicator function.}
\end{itemize}

To ensure heterogeneous differential privacy, we use a modified logistic loss. Therefore, our algorithm is as follows:
\begin{algorithm}[H]\caption{Minimizing Heterogeneous Differential Privacy}\label{alg:basic}
\begin{enumerate}
    \item Choose $(\boldsymbol{a}, \eta) \in \mathbb{F}$, where the constraint set is $\mathbb{F} 
 = \{(\boldsymbol{a}, \eta): \eta > 0, \ \sum_{i}^{} a_i = 1, \ 0 \leq a_i \leq k/m, \ a_i \eta \leq \epsilon_i \ \forall i\}$, where $k$ is a fixed constant.
    \item Pick, $\|\boldsymbol{b}\| \sim \Gamma(n,1)$ and direction chosen uniformly at random. Let $\boldsymbol{b'} = \frac{2\boldsymbol{b}}{\eta}$.\footnote{$||\boldsymbol{b'}||$ is drawn from Gamma Distribution}
    \item Given a dataset $D$ and differential privacies $\boldsymbol{\epsilon}$, compute $\hat{\boldsymbol{w}} = \mathrm{argmin}_{\boldsymbol{w}} \hat{\mathcal{L}}_m(D,\boldsymbol{w},\boldsymbol{a},\eta)$, where 
    \begin{align}\label{eq:L-hat}
    \hat{\mathcal{L}}_m(D,\boldsymbol{w},\boldsymbol{a},\eta) &= \sum_{i=1}^{m} a_i \log(1+e^{-y^i \cdot \boldsymbol{w}^T \boldsymbol{x}^i})\cr
    &\qquad+  \boldsymbol{b'}^T \boldsymbol{w} + \frac{\Lambda}{2} ||\boldsymbol{w}||^2,
    \end{align}
    for some $\Lambda > 0$. Output $\hat{\boldsymbol{w}}$.
\end{enumerate}
\end{algorithm}
\begin{proposition}\label{algoproof}
     For any $(\boldsymbol{a},\eta) \in \mathbb{F}$, the output of Algorithm \ref{alg:basic}, denoted by $\hat{w}$, preserves $\boldsymbol{\epsilon} + \Delta$ differential privacy, where $\Delta =  2 \log\big(1 + \frac{k}{m \Lambda}\big).$
\end{proposition}

\begin{remark}
Note that in most machine learning models, $m$ is large enough such that $m \Lambda >> 1$ and thus the term $\Delta$ is much smaller than the differential privacy guarantees used in practice. Therefore, for brewity, we consider $(\epsilon_i + \Delta) \approx \epsilon_i$ for further analysis.
\end{remark}

We note that multiple choices of $\boldsymbol{a}$ and $\eta$ can be selected to satisfy the privacy requirements. For example, one can choose $a_i = 1/m$ and $\eta = m \min_{i} \epsilon_i$. However, such a choice is unable to exploit the fact that we need to protect some data points more than others. Besides, our numerical experiments demonstrate that solving 
$\min_{\boldsymbol{w}} \hat{\mathcal{L}}(D,\boldsymbol{w},\boldsymbol{a},\eta)$ over the feasible set of $(\boldsymbol{a}, \eta)\in \mathbb{F}$ does not provide good results. 
Therefore, to understand how to choose $(\boldsymbol{a}, \eta)$, we appeal to statistical learning theory to get an upper bound on $\mathbb{E}[\mathbb{I}_{\{sign(\boldsymbol{w}^T \boldsymbol{x}) \neq y\}}]$ in terms of $\hat{\mathcal{L}}(D,\boldsymbol{w},\boldsymbol{a},\eta)$ with the true loss. This leads to following result.
\begin{theorem}
\label{gen loss}
Given a classification task, let $D$ be the dataset from $m$ users with iid datapoints such that $(\boldsymbol{x}^i,y^i) \sim \mathcal{D}$ and $\|\boldsymbol{x}^i\| \leq 1$. Further, consider that users have differential privacy requirements $\boldsymbol{\epsilon} = (\epsilon_i)_{i=1}^m \in \mathbb{R}_{+}^m$. Also, let $\mathcal{L}_m(D,\boldsymbol{c};\boldsymbol{\epsilon},\boldsymbol{\boldsymbol{w}}) = \mathbb{E}_{(\boldsymbol{x},y)\sim \mathcal{D}}[\mathbb{I}_{\{sign(\boldsymbol{w}^T \boldsymbol{x}) \neq y\}}]$ be misclassification loss and $\hat{\mathcal{L}}_m(D,\boldsymbol{w},\boldsymbol{a},\eta)$ be defined as in Eq. \eqref{eq:L-hat}. Then, for every choice of $\boldsymbol{w}$ such that $\|\boldsymbol{w}\| \leq \beta$ for some $\beta > 0$, $\boldsymbol{\epsilon} \in \mathbb{R}^m$, and $(\boldsymbol{a}, \eta) \in \mathbb{F}$, the following holds with probability at least $(1-\delta-\delta')$
 \begin{align}
 \label{derived-loss-1}
 \mathbb{E}_{(\boldsymbol{x},y)\sim \mathcal{D}}[&\mathbb{I}_{\{sign(\boldsymbol{w}^T \boldsymbol{x}) \neq y\}}]\leq \hat{\mathcal{L}}_m(D,\boldsymbol{w},\boldsymbol{a},\eta)\cr
 &+ \mu(\delta,\beta) \|\boldsymbol{a}\| +  \sigma(\delta,\delta',\beta) \Big(\frac{1}{\eta}\Big), 
\end{align}
where \begin{align}\nonumber &\mu(\delta,\beta) = \big(\frac{3  \ln \frac{1}{\delta}}{\sqrt{2}}\big) \log(1+e^{\beta}) + \frac{\beta}{\ln2},\cr &\sigma(\delta,\delta',\beta) = \big(\frac{6  \ln \frac{1}{\delta}}{\sqrt{2}} + 1\big) \big(2\beta v^{-1}(\delta')\big), \cr
&v(t) = \sum_{i=0}^{n-1} \frac{(t/2)^i}{i!} e^{-\frac{t}{2}}.
\end{align}
\hfill $\blacksquare$
\end{theorem}

Since, $\mu,\sigma$ depend on $\delta,\delta'$ which are themselves design specific, we consider $\mu, \sigma$ as hyperparameters whose choice will be explained in a later section. We denote the sum $\mu\|\boldsymbol{a}\| + \sigma/\eta$ as the excess risk. Using the above bound for generalization error, we note that for optimum performance of the logistic regression algorithm with heterogenous differential privacy guarantees, the excess risk should be minimized wrt $(\boldsymbol{a}, \eta) \in \mathbb{F}$ along with minimizing the empirical error $\hat{\mathcal{L}}(D,\boldsymbol{w},\boldsymbol{a},\eta)$. In the next section, we will use this result to formulate and analyze our algorithm for the mechanism design problem.

\section{Mechanism Design Problem}
\label{section: mech design}
We will now apply our logistic regression result to consider the mechanism design problem which aims to solve the privacy-utility tradeoff in differential privacy. We consider a platform (buyer) interested in collecting data from privacy-sensitive users (sellers) to build a logistic regression model. Further, sellers may have different costs associated with the privacy lost by sharing their data, i.e., they may have different privacy sensitivities. Therefore, the platform buys data from sellers in exchange for payments and provides them with differential privacy guarantees. The differential privacy guarantees are determined by optimizing an objective consisting of the misclassification error and the payments. More specifically, our problem has the following components:
\begin{figure}[H]
    \centering
    \includegraphics[scale = 0.5]{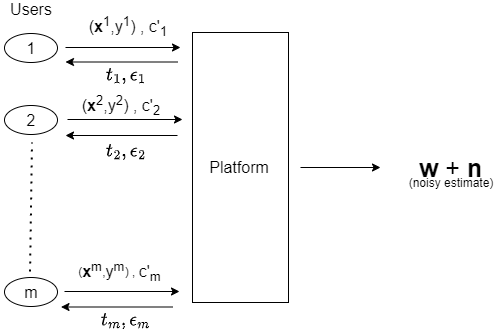}
        \caption{Interaction between sellers and the platform }
\end{figure}
\begin{itemize}
    \item We have a set of $m$ sellers, with seller $i$ having iid data $z^i = (\boldsymbol{x}^i,y^i) \sim \mathcal{D}$, with $n$-dimensional input $\boldsymbol{x}$ and output $y$ where $y^i\in\{+1,-1\}$. We let $D=\{(\boldsymbol{x}^1,y^1),\ldots,(\boldsymbol{x}^m,y^m)\}$ denote the dataset.
    \item We model the cost that the sellers incur due to loss of privacy using privacy sensitivities $c_i\geq0, i=1,\ldots,m$. In other words, if a seller is provided a differential privacy guarantee of $\epsilon_i,$ then the seller incurs a cost of $c_i\cdot \epsilon_i$.
    \item Sellers can potentially lie about their privacy sensitivity to get an advantage. Therefore, we denote the reported privacy sensitivity of seller $i$ by $c'_i$. The mechanism will be designed so that sellers report truthfully, i.e., $c_i'=c_i.$
    \item As is standard in the mechanism design literature, we assume that sellers' cost $c_i$ are drawn iid from a common probability density function $f_C(\cdot)$, which is common knowledge. Moreover, we assume that sellers cannot lie about their data. This assumption is valid in scenarios such as healthcare data, where patient information is already within the possession of the hospital. In this context, sellers merely need to grant permission to the hospital (a trusted authority) to utilize their data, specifying their privacy sensitivities in the process.
    \item The mechanism proceeds as follows: i) First, sellers share their privacy sensitivities $\boldsymbol{c}$ with the buyer. ii) Based on $\boldsymbol{c}$, the buyer provides payments $t_i$ and guarantees a privacy $\epsilon_i$ to each seller $i$. iii) After the sellers receive their respective payments, they release their data to the buyer, who uses it to train their ML model which must adhere to the promised privacy guarantees $\boldsymbol{\epsilon}$.\footnote{$\boldsymbol{c} = [c_1, c_2, \ldots, c_m].$ Same notation is used in writing $\boldsymbol{\epsilon}, \boldsymbol{c'}, \boldsymbol{t}$.}
    \item Based on the privacy loss and the payment received, the cost function of seller $i$ with privacy sensitivity $c_i$, reported privacy sensitivity $c'_i$, and data point $z^i = (\boldsymbol{x}^i, y^i)$, is given by 
    \begin{equation}
    \label{seller-cost1}\mbox{COST}(c_i,\boldsymbol{c}_{-i},c'_i,\boldsymbol{c}_{-i}; \epsilon_i, t_i) =    c_i \cdot\epsilon_i - t_i.
\end{equation}
\end{itemize}

\subsection{Buyer's Objective}
To design the mechanism, we next state the buyer's goal.
\begin{itemize}
    \item  The buyer learns an ML model $\theta(D,\boldsymbol{c'})$ from  dataset $D$ and the vector of privacy sensitivities $\boldsymbol{c'}$, and computes a payment $t_i(D,\boldsymbol{c'})$ to seller $i$ while guaranteeing a privacy level $\epsilon_i(D,\boldsymbol{c'})$ to each seller $i.$ To do this, the buyer optimizes a combination of the test loss incurred by the ML model $\mathcal{L}_m(D,\boldsymbol{c'};\boldsymbol{\epsilon,\theta})$ and the payments $t_i(D,\boldsymbol{c'})$. The overall objective of the buyer is to minimize the cost function 
    \begin{align}\label{main-loss}   &\qquad \qquad \min_{\boldsymbol{\epsilon}, \boldsymbol{\theta}, \boldsymbol{t} } \mathbb{E}_{\boldsymbol{c}}[\mathcal{J}_m(D,\boldsymbol{c'},\boldsymbol{\epsilon},\boldsymbol{\theta},\boldsymbol{t})] , \\
     &\mathcal{J}_m(D,\boldsymbol{c'},\boldsymbol{\epsilon},\boldsymbol{\theta},\boldsymbol{t}) = 
    \mathcal{L}_m(D,\boldsymbol{c'};\boldsymbol{\epsilon,\theta})+\gamma\sum_{i=1}^m t_i(D,\boldsymbol{c'}), \nonumber 
    \end{align}
    where $\gamma$ is a hyperparameter that adjusts the platform's priority to get a better predictor or reduce payments.

    Note that since the buyer also guarantees the privacy of the sellers' data, the optimal model $\theta(D,\boldsymbol{c}')$ is required to be consistent with the privacy guarantees $\boldsymbol{\epsilon}(D,\boldsymbol{c}').$
    \item The buyer is also interested in ensuring each seller is incentivized to report their privacy sensitivity truthfully. To that end, the \emph{incentive compatibility} (IC) property imposes that no seller would misrepresent their privacy sensitivity if others report truthfully, i.e.,
\begin{align}
\label{Eq: IC}
&\mbox{COST}(c_i,\boldsymbol{c}_{-i},c_i,\boldsymbol{c}_{-i}; \epsilon_i, t_i) \cr
&\leq \mbox{COST}(c_i,\boldsymbol{c}_{-i},c'_i,\boldsymbol{c}_{-i}; \epsilon_i, t_i) \quad \forall i, c'_i, \boldsymbol{c}.
\end{align}
\item Moreover, the buyer wants to ensure that sellers are incentivized to participate. Thus, the \emph{indivitual rationality} (IR) property imposes the constraint that the platform does not make sellers worse off by participating in the mechanism, i.e.,
\begin{equation}
\label{Eq: IR}
\mbox{COST}(c_i,\boldsymbol{c}_{-i},c'_i,\boldsymbol{c}_{-i}; \epsilon_i, t_i) \leq 0 \quad \forall i, c'_i, \boldsymbol{c}.
\end{equation}
\end{itemize}
Therefore, the order of operations of our mechanism can be summarized as follows. 
\begin{enumerate}
   \item The sellers provide the platform/buyer with their privacy sensitivity $c'_i$ and their data. 
    \item The buyer announces that in exchange for the data, it will pay according to a payment rule that must satisfy the IC and IR constraints. 
    \item Next, the buyer proposes privacy levels $\boldsymbol{\epsilon}(D,\boldsymbol{c'})$ and payments $\boldsymbol{t}(D,\boldsymbol{c'})$ that will be guaranteed to each seller in exchange for their data.
    \item The buyer then obtains an ML model $\theta(D,\boldsymbol{c'})$ which is revealed to the public and is therefore required to be consistent with the proposed privacy guarantees $\boldsymbol{\epsilon}(D,\boldsymbol{c'})$.
\end{enumerate}

Now, to proceed with deriving an optimal solution to the mechanism, we first show that if $\epsilon_i(D,\boldsymbol{c'})$ is the privacy guarantee provided to seller $i$, then using the IC and IR constraints, we can replace the payments $t_i(D,\boldsymbol{c'})$ in the objective function by $\psi(c_i) \epsilon_i (D,\boldsymbol{c'})$, where $\psi(c) =  c + F_C(c)/f_C(c)$.\footnote{$F_C(c_i)$ and $f_C(c_i)$ denote the values of CDF and PDF functions for privacy sensitivities at $c_i$, respectively.}
Furthermore, the IC constraint incentivizes sellers to be truthful, and henceforth, we can replace $\boldsymbol{c'}$ with $\boldsymbol{c}$. We state our result along with the regularity assumption formally as follows
\begin{assumption}
\label{assumption: virtual cost}
The virtual cost $\psi(c) = c + \frac{F_C(c)}{f_C(c)}$ is a strictly increasing function of $c$.
\end{assumption}
\begin{lemma}
\label{mech design}
Assume that $c_{i}$ is drawn from a known pdf $f_C(\cdot)$ with cdf $F_C(\cdot)$, such that the virtual cost satisfies Assumption \ref{assumption: virtual cost}. Then given a mechanism design problem with the sellers' costs given by Eq. (\ref{seller-cost1}), if $\boldsymbol{c}$ denotes the sellers' privacy sensitivities and $\boldsymbol{\epsilon}(D,\boldsymbol{c})$ denote privacy guarantees promised to the sellers, the payment model
\begin{equation}
\label{Eq: payments}
    t_i(D,\boldsymbol{c}) = -\int_{c_i}^{\infty} z\frac{d}{dz} \epsilon_i(D,\boldsymbol{c}_{-i},z) dz.
\end{equation}
satisfies the IC and IR constraints. Further, we can write
    \begin{equation}
   \mathbb{E}_{\boldsymbol{c}}\big[\sum_{i=1}^{m} t_i(D,\boldsymbol{c}) \big] =   \mathbb{E}_{\boldsymbol{c}}\big[\sum_{i=1}^{m} \epsilon_i(D,\boldsymbol{c}) \psi(c_i) \big].
\end{equation}
\hfill $\blacksquare$
\end{lemma}

\smallskip
Using Lemma \ref{mech design}, we can write the payments in terms of differential privacy guarantees $\boldsymbol{\epsilon}$. This helps to separate the problem into two parts, namely, designing the payments to be compatible with the IC and IR constraints and the learning problem to optimize Eq. (\ref{main-loss}), which obtains the optimal model weights $\boldsymbol{w}$ and the privacy guarantees $\boldsymbol{\epsilon}$. Therefore, on applying Lemma \ref{mech design} to the mechanism design formulation, we get the following result. 
\begin{theorem}
\label{thm: gen mechanism result}
Consider $(\boldsymbol{\epsilon}^*, \boldsymbol{\theta}^*, \boldsymbol{t}^*)$ to be the optimal solution of the mechanism design problem with the Buyer's objective given by Eq. \eqref{main-loss} and the payments constrained to satisfy the IC and IR constraints, i.e., Eq. \eqref{Eq: IC} and Eq. \eqref{Eq: IR}. Then the optimal solution satisfies 
\begin{equation}\nonumber
t^*_i(D,\boldsymbol{c}) = -\int_{c_i}^{\infty} z\frac{d}{dz} \epsilon^*_i(D,\boldsymbol{c}_{-i},z) dz,
\end{equation}
where $(\boldsymbol{\theta}^*,\boldsymbol{\epsilon}^*)$ is the solution to the optimization problem
\begin{equation} 
\label{Eq: buyers obj2}
(\boldsymbol{\theta}^*, \boldsymbol{\epsilon}^*) = \arg\min_{\theta,\boldsymbol{\epsilon}} \bigg[\mathcal{L}_m(D,\boldsymbol{c};\boldsymbol{\epsilon},\boldsymbol{\theta}) + \gamma \cdot \sum_{i=1}^{m}\psi(c_{i})\epsilon_{i}\bigg], 
\end{equation}
subject to the constraint that $\boldsymbol{\theta}^*$ is $\boldsymbol{\epsilon}^*$-differentially private.\hfill $\blacksquare$
\end{theorem}


We make the following remarks regarding Thm \ref{thm: gen mechanism result} and possible approaches to solve Eq. \eqref{Eq: buyers obj2}.
\begin{enumerate}
    \item Note that the payment mechanism decided according to Lemma \ref{mech design} does not depend on the choice of loss function. So the above theorem holds for other problems, not just logistic regression.
    \item Since in general there does not exist a mathematical representation for the misclassification error  $\mathcal{L}_m(D,\boldsymbol{c};\boldsymbol{\epsilon},\boldsymbol{\theta}) = \mathbb{E}_{(\boldsymbol{x},y)\sim \mathcal{D}}[\mathbb{I}_{\{sign(\boldsymbol{w}^T \boldsymbol{x}) \neq y\}}]$, we cannot directly optimize Eq. \eqref{Eq: buyers obj2}. Therefore, we consider the upper bound in Thm \ref{gen loss} as a surrogate loss. Appendix \ref{sec: data dependent pay} provides an algorithm to minimize the surrogate loss while ensuring a unique optimal solution.
    \item One issue with the mechanism in the above theorem is that the payments depend on the data and thus, may leak privacy. While the payments are only revealed to the sellers (and not to the public), if a seller is inquisitive, they may potentially learn about other sellers' data. This may be acceptable if one assumes that individual sellers do not have the computational power to cause privacy leakage. 
\end{enumerate}
To address the issue of privacy leakage for a general case, in the next section, we design a mechanism that does not use the data to decide payments and is also asymptotically optimal in the limit when the number of users becomes large. Such a mechanism may also be more attractive to sellers since they do not have to share their data before payments are made.

\section{Solving the Mechanism Design Problem}
\label{sec: algo}
To mitigate privacy loss through payments to sellers and to allow sellers to provide data after payments, the key idea in this section is to replace the test loss with excess risk, i.e., test loss minus training loss and separately solve for the privacy guarantees and model weights. As no exact expressions are known for the excess risk, we work with known upper bounds as derived in Thm \ref{gen loss}. Our proposed mechanism design algorithm works as follows: 

\begin{enumerate}
    \item First we obtain the privacy guarantees $\epsilon_i$ in terms of privacy sensitivities $\boldsymbol{c}$ that minimizes a weighted sum of payments and the excess risk. More precisely, the algorithm computes $\boldsymbol{a}, \eta$ by optimizing
    \begin{align}
    \label{Eq: utility privacy tradeoff}
       &\qquad \quad \min_{\boldsymbol{a},\eta, \boldsymbol{\epsilon}} \ \mathbb{L}_m(\boldsymbol{c},\boldsymbol{\epsilon}), \nonumber \\
       & \mathbb{L}_m(\boldsymbol{c},\boldsymbol{\epsilon}) = \mu \|\boldsymbol{a}\| + \frac{\sigma}{\eta} + \gamma  \sum_{i=1}^{m} \epsilon_i \psi(c_i).
       \end{align}
    subject to $ \eta > 0$, $\boldsymbol{a} \geq 0$, $a_i \eta \leq \epsilon_i$, $a_i \leq k/m \ \forall i$, and $\sum_i a_i = 1$. Here, $\mu, \sigma$, and $\gamma$ are the hyperparameters defined as before. First, we observe that Eq. (\ref{Eq: utility privacy tradeoff}) is optimized wrt $\boldsymbol{\epsilon}$ when $\boldsymbol{\epsilon} = \boldsymbol{a} \eta$ thereby reducing the problem to an optimization problem over $(\boldsymbol{a}, \eta).$
    Next, we note that Eq. (\ref{Eq: utility privacy tradeoff}) is convex with respect to $\boldsymbol{a}$ when $\eta$ is fixed. Therefore to solve Eq. (\ref{Eq: utility privacy tradeoff}), we begin by fixing $\eta$ and optimizing the equation wrt $\boldsymbol{a}$ using projected gradient descent. Additionally, note that $\eta = \sum_i \epsilon_i$. Since $\boldsymbol{\epsilon}$ cannot be excessively high in practical scenarios, we can bound the range of $\eta \in [0,L]$ for some constant $L > 0$. We then discretize this interval to the desired precision and perform a line search over all possible values of $\eta$.
    Denote the optimal values to be $(\hat{\boldsymbol{a}}, \hat{\eta},\hat{\boldsymbol{\epsilon}})$. Also, we will denote $\mathbb{L}_m(\boldsymbol{c},\boldsymbol{\epsilon})$ as the proxy loss.
    \item  Using the estimated values $\hat{\boldsymbol{\epsilon}} = \hat{\boldsymbol{a}}\hat{\eta}$, we design a differentially private logistic regression estimator $\hat{\boldsymbol{w}}$ using 
     \begin{equation}
    \label{Eq: diff privacy eqn}
        \min_{\boldsymbol{w}} \sum_{i=1}^{m} \hat{a}_i \log(1+e^{-y^i \cdot \boldsymbol{w}^T \boldsymbol{x}^i}) +  \frac{2\boldsymbol{b}^T \boldsymbol{w}}{\hat{\eta}} + \frac{\Lambda}{2} \|\boldsymbol{w}\|^2,
    \end{equation}
    where $\boldsymbol{b}$ is a random vector whose norm follows the Gamma distribution $\|\boldsymbol{b}\| \sim \Gamma(n,1),$ and the direction of $\boldsymbol{b}$ is chosen uniformly at random. We also note that using Proposition \ref{algoproof} we can show that $\hat{\boldsymbol{w}}$ ensures $\hat{\boldsymbol{\epsilon}}$-differential privacy. Further, since the payments are independent of the training data, they also do not leak any privacy.
\end{enumerate}    

We refer to Algorithm \ref{alg:modified-leak} for a summary of the steps of our mechanism design algorithm
\begin{algorithm}[H]\caption{
Mechanism Design}\label{alg:modified-leak}
\textbf{Step 1:} Calculate $\hat{\boldsymbol{\epsilon}}$ by minimizing the sum of excess risk and payments
\begin{equation}\nonumber
(\hat{\boldsymbol{a}},\hat{\eta})=\arg\min_{\boldsymbol{a},\eta} \Big\{\mu \|\boldsymbol{a}\| + \frac{\sigma}{\eta} + \gamma   \sum_{i=1}^{m} \eta a_i \psi(c_i)\Big\},
  \end{equation}
  where $\hat{\boldsymbol{\epsilon}} = \hat{\boldsymbol{a}} \hat{\eta}$.\\
\textbf{Step 2:} Use the payment identity to calculate the payments $\hat{t}_i(\boldsymbol{c})$ from the obtained privacy guarantees $\hat{\boldsymbol{\epsilon}}$, i.e.,
\begin{equation}\nonumber
     \hat{t}_i(\boldsymbol{c}) = -\int_{c_i}^{\infty} z\frac{d}{dz} \hat{\epsilon}_i(\boldsymbol{c}_{-i},z) dz.
\end{equation}
\textbf{Step 3:} Solve for $\hat{\boldsymbol{w}}$ to obtain the optimal weights consistent with the differential privacy guarantees $\hat{\boldsymbol{\epsilon}}$:
 \begin{equation}\nonumber
     \min_{\boldsymbol{w}} \sum_{i=1}^{m} \hat{a}_i \log(1+e^{-y^i \cdot \boldsymbol{w}^T \boldsymbol{x}^i}) +  \frac{2\boldsymbol{b}^T \boldsymbol{w}}{\hat{\eta}} + \frac{\Lambda}{2} \|\boldsymbol{w}\|^2.
 \end{equation}
\end{algorithm}
\begin{remark}
The constraint $a_i \leq k/m$ for some $k > 0$ indirectly imposes the condition that $\epsilon_i$ is upper-bounded by a finite quantity. Since $\eta = \sum_{i} \epsilon_i$, we can write $\eta = m\epsilon_{avg}$, which together with $\epsilon_i = a_i \eta$ implies $\epsilon_i \leq k \epsilon_{avg}$. Therefore, the upper bound constraint on $a_i$ means that $\epsilon_i$ is upper-bounded. In other words, using our formulation we are also able to incorporate practical constraints such as sellers' unwillingness to tolerate more than a certain amount of privacy loss even if they are paid generously.
\end{remark}
\begin{remark}
In addition to $(\boldsymbol{w}, \boldsymbol{a}, \eta)$, our objective function contains a set of hyperparameters $\{\Lambda, \mu, \sigma, \gamma\}$. Therefore, we require a validation dataset to compare different values of the hyperparameters. Since hyperparameters are chosen based on the validation set, if the validation dataset is private, it may violate differential privacy guarantees. \cite{kamalika} provides a detailed discussion on ensuring differential privacy using validation data. Referring to their work, we assume the existence of a small publicly available dataset that can be used for validation. This assumption ensures that differential privacy guarantees are not violated.
\end{remark}
\subsection{Discussion}
We make the following remarks on our solution:
\begin{itemize}

\item The payment mechanism is independent of $\mathcal{L}_m(D,\boldsymbol{c};\boldsymbol{\epsilon},\boldsymbol{\boldsymbol{w}})$.
Therefore, if one uses a different algorithm for the learning part of the mechanism, i.e., Eq. (\ref{Eq: buyers obj2}) it would not affect the behavior of sellers, i.e., sellers will still be incentivized to be truthful, and willing to participate in the mechanism.

\item Since the payment mechanism does not depend on the choice of function $\mathbb{L}_m(D,\boldsymbol{c};\boldsymbol{\epsilon},\boldsymbol{\boldsymbol{w}})$, designing payment mechanism and solving objective function can be treated as two separate problems. Thus, any such mechanism design problem can be decoupled into separate problems.

\item Finally, we can see that our algorithm can be used to solve the logistic regression problem with heterogeneous privacy guarantee requirements. Previous works, such as \cite{kamalika}, solves the problem in the case when it is assumed that all users have the same differential privacy requirements. In our work, Eq. (\ref{derived-loss-1}) extends it to the case when users are allowed to have different privacy requirements.
\end{itemize}
\subsection{Performance Analysis}\label{subsec:conditions}
Recall the discussion below Eq. (\ref{Eq: utility privacy tradeoff}), where we showed that optimizing $\mathbb{L}_m(\boldsymbol{\epsilon},\boldsymbol{c})$ with respect to $\boldsymbol{a},\eta,\boldsymbol{\epsilon}$ shows that the optimal solution should satisfy $\boldsymbol{\epsilon} = \boldsymbol{a} \eta$ and $\eta=\sum_i\epsilon_i.$ 
Thus, by substituting for $\boldsymbol{a}$, we can treat the problem as one of optimizing over $\boldsymbol{\epsilon}$ and $\eta$ under the constraint $\eta=\sum_i\epsilon_i.$
Using a Lagrange multiplier $\lambda$, we obtain the Lagrangian function as
 \begin{equation}\nonumber
      \frac{\mu \|\boldsymbol{\epsilon}\|}{\eta} + \frac{\sigma}{\eta} + \gamma \sum_{i}^{m} \epsilon_i \psi(c_i)+ \lambda(\eta-\sum_i \epsilon_i),
 \end{equation}
 such that $\epsilon_i \geq 0, \eta \geq 0$. Therefore, using the first-order necessary conditions, we obtain
 \begin{align}\nonumber
     &\epsilon_i = \big(\lambda-\gamma\psi(c_i)\big)^{+} \frac{\|\boldsymbol{\epsilon}\|\eta}{\mu}, \cr
     &\lambda = \frac{\sigma + \mu \|\boldsymbol{\epsilon}\|}{\eta^2},
 \end{align}
 where $(a)^+=\max\{0,a\}$. Using these equations, we can write
\begin{equation}
\label{Eq: eta eqn}
    \epsilon_i = \frac{\mu(\lambda-\gamma\psi(c_i))^{+}}{\|(\lambda-\gamma\psi(\boldsymbol{c}))^{+}\| \ \sum_i (\lambda-\gamma\psi(c_i))^{+}},
\end{equation}
with $\lambda$ such that
\begin{equation}
\label{Eq:lambda}
    \lambda - \frac{\sigma}{\mu^2} \|(\lambda-\gamma\psi(\boldsymbol{c}))^{+}\|^2 - \frac{\|(\lambda-\gamma\psi(\boldsymbol{c}))^{+}\|^2}{\sum_i (\lambda-\gamma\psi(c_i))^{+}} = 0.
\end{equation}

While solving the above set of equations in closed-form is difficult, in Lemma \ref{thm: offline perf} in the appendix, we show that, with high probability, $\mathbb{L}_m(\boldsymbol{c},\hat{\boldsymbol{\epsilon}}) = \mathcal{O}(m^{-1/4})$  under the following assumption, which essentially states that there must be non-trivial probability mass around the origin in the distribution of $c$.
\begin{assumption}
\label{assumption: pdf}
Let $f_{C}(\cdot)$ be the probability density function corresponding to $c$. There exists $c_1 > 0$ such that $0 < f_{C}(c) < \infty \ $for$ \ c \in [0,c_1].$\footnote{Similar results also hold under a weaker assumption such as $\frac{d^n}{dc^n} f_C(c)\neq 0$ in an interval around 0 for some $n$.}
\end{assumption}
Next, we assume that the data set is separable to establish our main result on the buyer's objective.
\begin{assumption}
\label{assumption: sep}
   The data set is linearly separable, that is, there exists a $\boldsymbol{w^{*}}$ with $\|\boldsymbol{w}^*\| = 1$ such that $ \boldsymbol{w^{*}}^T\boldsymbol{x^i}y^i \geq \rho \ \forall i$, for some $\rho > 0$.
\end{assumption}
\begin{theorem}
\label{thm: asymptotic}
Assume that the dataset $D$ and the pdf $f_{\Psi}(.)$ satisfy Assumptions \ref{assumption: pdf} and \ref{assumption: sep}. Furthermore, let $(\boldsymbol{\hat{w}}(D,\boldsymbol{c}), \boldsymbol{\hat{\epsilon}}(\boldsymbol{c}))$ be the solution obtained using Algorithm \ref{alg:modified-leak} such that $\|\boldsymbol{\hat{w}}\| = \beta$. Additionally, the payments $\boldsymbol{\hat{t}}(\boldsymbol{c})$ are chosen using the payment rule Eq. (\ref{Eq: payments}). Then, for every $\nu > 0$, the buyers objective (see Eq. (\ref{main-loss})) is bounded as follows
\begin{align}\label{eq:asym-upper-bound}
\mathbb{P}\Big(\mathcal{J}_m(D,\boldsymbol{c},\boldsymbol{\hat{\epsilon}},\boldsymbol{\hat{w}},\boldsymbol{\hat{t}})&\geq \log(1+e^{-\rho\beta})  + \nu + \mathrm{o}(1) \Big) \nonumber \\
   & \leq \mathcal{O}(e^{-\nu \frac{m^{1/4}}{2\beta}}),
\end{align}
where $\mathrm{o}(1)$ is a term that goes to zero as $m\rightarrow\infty.$
\hfill $\blacksquare$
\end{theorem}

The above theorem shows that one can make the objective $\mathcal{J}_m$ small with high probability by choosing $\beta$ sufficiently large, $\nu$ sufficiently small, and $m$ sufficiently large.
\smallskip
\section{Extension to Online Setting}
\label{section: online setting}
Next, we turn our attention to another variant of our mechanism design problem: the online version, where sellers arrive sequentially at the platform. Specifically, consider a setting where sellers arrive one by one to the platform, each reporting their privacy sensitivity upon arrival. Additionally, the sellers could be impatient/uncomfortable sharing their data without receiving an upfront payment. This imposes an additional constraint on the buyer, who must make immediate decisions regarding both the payments and the differential privacy guarantees as soon as a seller discloses their privacy sensitivity. More precisely, the online problem can be described as follows:
\begin{itemize}
    \item Sellers arrive sequentially to the buyer. Once the sellers arrive, they report their privacy sensitivities $c'_i$ to the buyer. We assume that the true privacy sensitivities $c_i$ are drawn from the same distribution with pdf $f_C(\cdot).$
    \item The platform then determines the differential privacy $\epsilon_i$ and the payment $t_i$ to the seller, which depends on $\epsilon_i$. At this point, if the seller agrees to this deal, it shares its data $(\boldsymbol{x}^i,y^i)$ with the platform.\footnote{To incentivize the seller to agree to the deal and also report their privacy sensitivities truthfully, we design the payments such that they satisfy IC and IR constraints.} Note that since the sellers arrive sequentially, $\epsilon_i$ can only be a function of past privacy sensitivities $\boldsymbol{c}_{1:i}=(c_1,\ldots,c_i)$.
    \item After $m$ sellers have interacted with the platform, it designs a classifier using the collected data consistent with differential privacy guarantees $\boldsymbol{\epsilon}$ provided to sellers.
\end{itemize}\begin{algorithm}\caption{An Online Algorithm for the Mechanism Design}\label{alg:online}
\textbf{Given:} $m$, probability density function $f_C(\cdot)$.\\
Using $f_C(\cdot)$, calculate the pdf of $\psi(\cdot)$ and denote it by $f_{\Psi}(\cdot)$. Set the cut-off price as $\tilde{\lambda} \gets \sqrt{\frac{\mu^2\gamma}{\sigma m f_{\Psi}(0)}}$.\\
\textbf{Transaction:} Whenever a new seller arrives, set
\begin{align}\nonumber
\tilde{\epsilon_i}=\begin{cases}
\frac{2\sqrt{3}\gamma^{3/2}\mu(\tilde{\lambda}-\gamma\psi(c_i))}{f^{3/2}_p(0)m^{3/2}\tilde{\lambda}^{7/2}} &\mbox{If}\ \ \gamma\psi(c_i) < \tilde{\lambda},\\
0 &\mbox{else}.
\end{cases}   
\end{align}
\textbf{Payments: }Calculate the payment $\tilde{t_i}$ using Eq. (\ref{Eq: payments}).\\
\textbf{Differentially Private Logistic Regression: }After $m$ sellers have interacted with the buyer, solve for the model weights $\tilde{\boldsymbol{w}}$ using Eq. (\ref{Eq: diff privacy eqn}).
\end{algorithm}
To solve the online mechanism design problem, we propose Algorithm \ref{alg:online} which is specifically designed to handle the sequential arrival of sellers, making real-time decisions on payments and privacy guarantees based on the reported privacy sensitivities. 
Additionally, in our algorithm, we assume that the platform continues to collect data until $m$ sellers have interacted with it. However, in practice, only a fraction $\xi\in (0,1]$ of the $m$ sellers may actually interact with the platform. The following result provides performance bounds for Algorithm \ref{alg:online} as well as quantifies the loss in performance due to the buyer's overestimation of the number of sellers
\begin{theorem}
\label{thm: perf bounds}
Assume that the dataset $D$ and the pdf $f_{\Psi}(\cdot)$ satisfy Assumptions \ref{assumption: pdf} and \ref{assumption: sep}. Additionally, let $\xi m$ be the total number of sellers which interact with the buyer sequentially for some $\xi \in (0,1].$ Furthermore, let $(\boldsymbol{\tilde{w}}(D,\boldsymbol{c}), \boldsymbol{\tilde{\epsilon}}(\boldsymbol{c}),\boldsymbol{\tilde{t}}(\boldsymbol{c}))$ be the solution obtained using Algorithm \ref{alg:online} such that $\|\boldsymbol{\tilde{w}}\| = \beta$. Then, for every $\nu > 0$, the buyer's objective is bounded as follows
\begin{align}\label{eq:asym-upper-bound2}
\mathbb{P}\Big(\mathcal{J}_m(D,\boldsymbol{c},\boldsymbol{\tilde{\epsilon}},\boldsymbol{\tilde{w}},\boldsymbol{\tilde{t}})&\geq \log(1+e^{-\rho\beta})  + \nu + \mathrm{o}(1) \Big) \nonumber \\
   & \leq \mathcal{O}(e^{-\nu \xi \frac{m^{1/4}}{2\beta}}),
\end{align}
where $\mathrm{o}(1)$ is a term that goes to zero as $m\rightarrow\infty.$
\hfill $\blacksquare$
Therefore, even for the online setting we achieve similar performance bounds as compared to the offline setting.
\end{theorem}.

\section{Numerical Results}
\label{sec: num}

In this section, we provide some numerical experiments to validate the efficacy of our devised algorithms. 

\subsection{Offline Algorithm}
\begin{figure}[H]    \centering
    \subfloat[\centering Misclassification error and payments]
   {{\includegraphics[width=3.9cm]{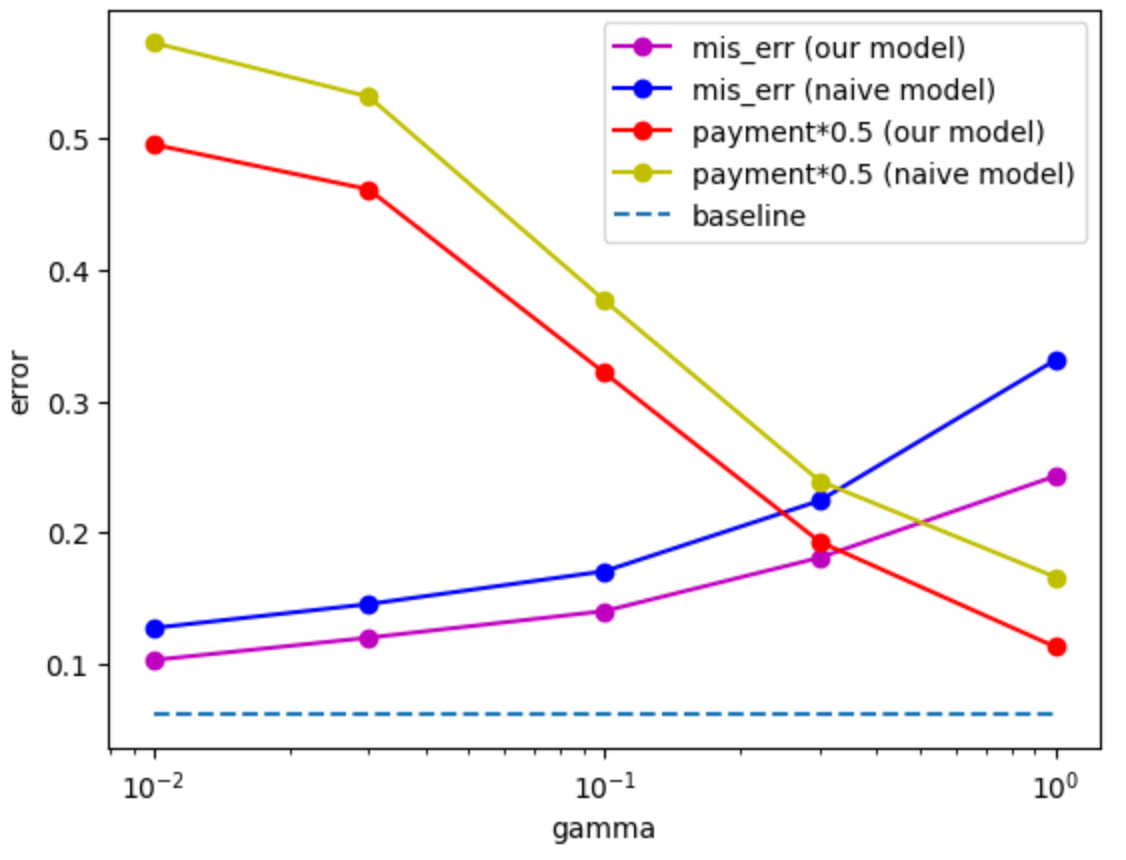} }}
    \qquad
    \subfloat[\centering Comparison of overall error]{{\includegraphics[width=3.9cm]{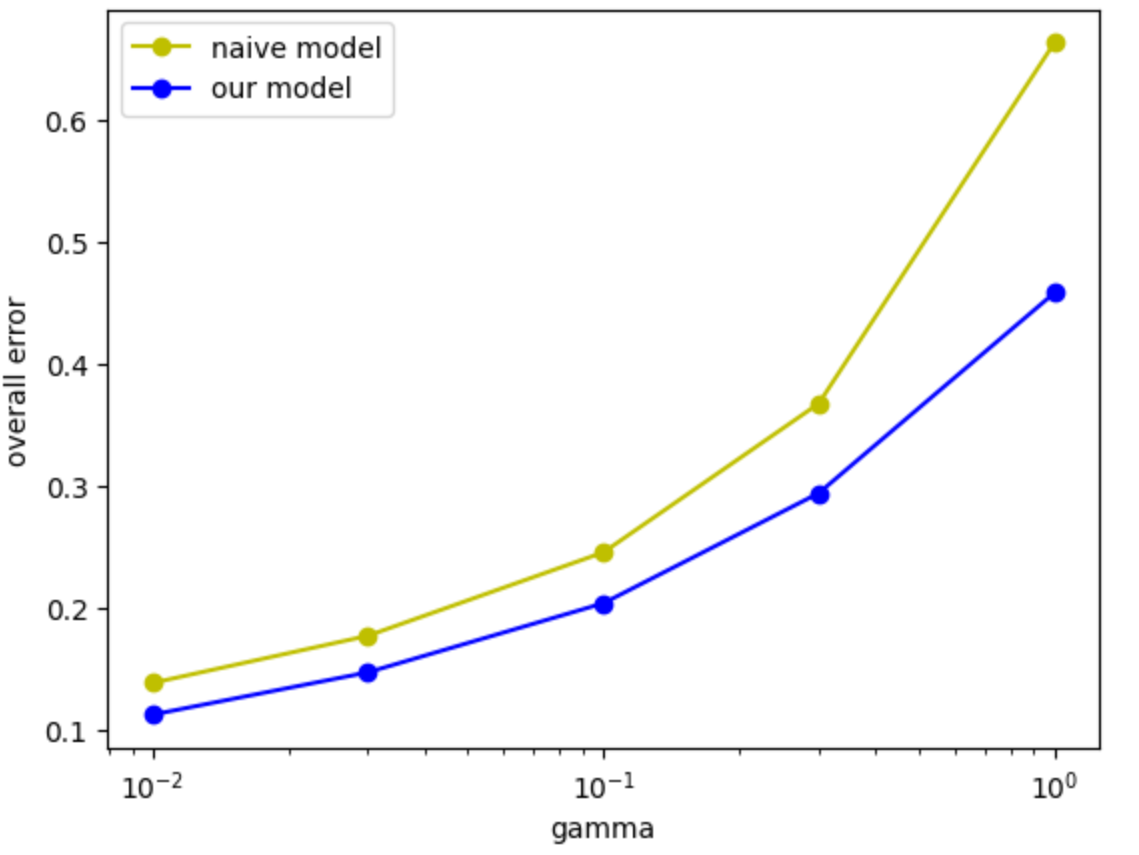} }}
    \caption{Mechanism Design Experiments }
    \label{fig: Mechanism Design Analysis}
\end{figure}

We first demonstrate how the offline mechanism design can be used in a practical application. Additionally, we also showcase the importance of considering the regularization terms ($\mu \|\boldsymbol{a}\|, \sigma/\eta$) in the loss function. \\
\textbf{Dataset and Specifications:} We perform our mechanism design approach on the Wisconsin Breast Cancer dataset \cite{data}. Furthermore, $\boldsymbol{c}$ is drawn from $\mathbb{U}[e^{-4},5e^{-4}]$ and $\psi(c)$ is calculated accordingly to be $\psi(c)=2c - e^{-4}$. \\
\textbf{Implementation:} The loss function Eq. (\ref{Eq: buyers obj2}) is optimized using the algorithm provided in Appendix \ref{sec: data dependent pay} with the hyperparameters $\{\Lambda, \mu, \sigma\}$ selected based on the validation data. Further, the corresponding misclassification error (misclassified samples/total samples) and payments are plotted for each $\gamma$ in Fig. \ref{fig: Mechanism Design Analysis}. The values are plotted by taking the mean over 15 different samples of the noise vector $\boldsymbol{b}$. Given that our approach is the first to consider the tradeoff between payments and model accuracy for ML models, there is a lack of existing methods in the literature for direct comparison. However, to showcase the increase in efficiency of our approach due to the addition of extra regularization terms ($\mu \|\boldsymbol{a}\|, \sigma/\eta$), we compare our results with a \textit{naive model} whose objective does not consider these terms, i.e., Eq. (\ref{Eq: buyers obj2}) with $\mu=\sigma=0$. Finally, all results are benchmarked with the baseline error, which is the misclassification error of the model in the absence of payments and differential privacy guarantees. Additionally, to evaluate the efficiency of all the methods, the overall error (misclassification error + $\gamma \times$payments) is plotted in Fig. \ref{fig: Mechanism Design Analysis}. It is important to note that additional experiments in the appendix provide further insight about the hyperparameters.\\
\textbf{Observations and Practical Usage:} As depicted in Fig. \ref{fig: Mechanism Design Analysis}, there is a tradeoff between misclassification loss and payments, with an increase in misclassification loss and a decrease in payments as $\gamma$ rises. Consequently, the platform can tailor $\gamma$ based on its requirements. For example, if the platform has a budget constraint, the platform can iteratively adjust $\gamma$ to obtain the optimal estimator within the given budget. Finally, from Fig. \ref{fig: Mechanism Design Analysis}, we see that the incorporation of regularization terms ($\mu \|\boldsymbol{a}\|, \sigma/\eta$)  in the model yields a more efficient mechanism with a lower overall error.

\vspace{-0.5cm}
\begin{figure}[H]    \centering
     \label{fig: proxy loss }
    \subfloat[\centering Online vs Offline Algorithm]
   {{\includegraphics[width=3.9cm]{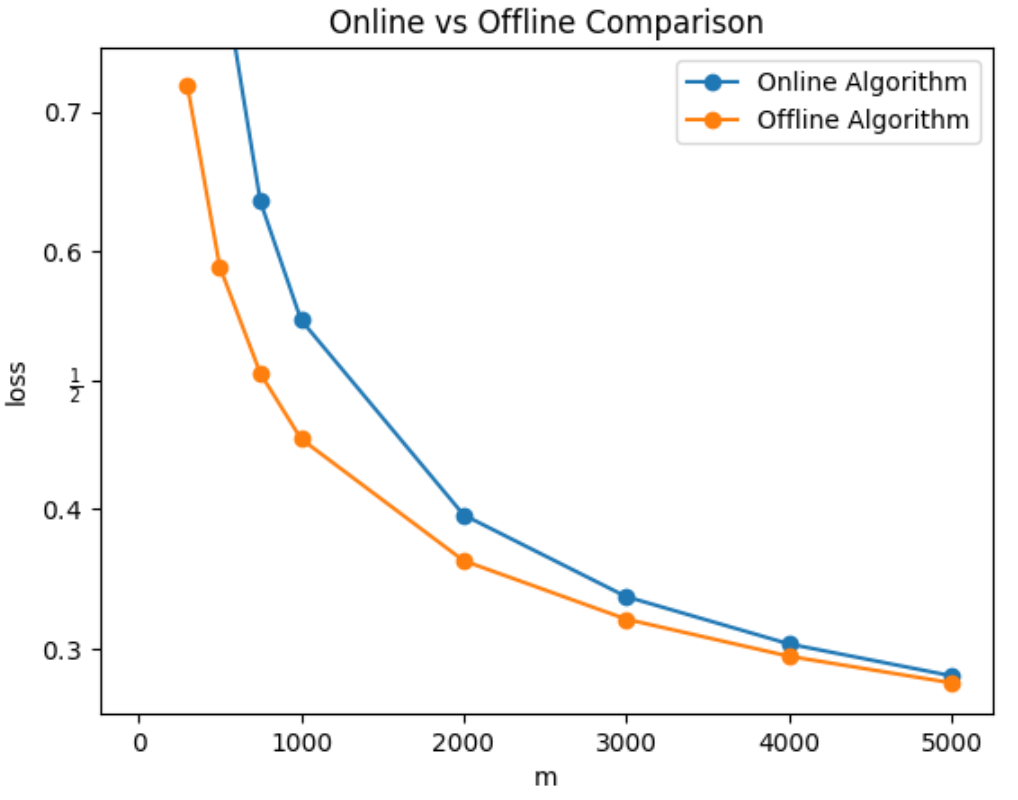} }}
    \qquad
    \label{fig: comp ratio}
    \subfloat[\centering Competitive Ratio]{{\includegraphics[width=3.9cm]{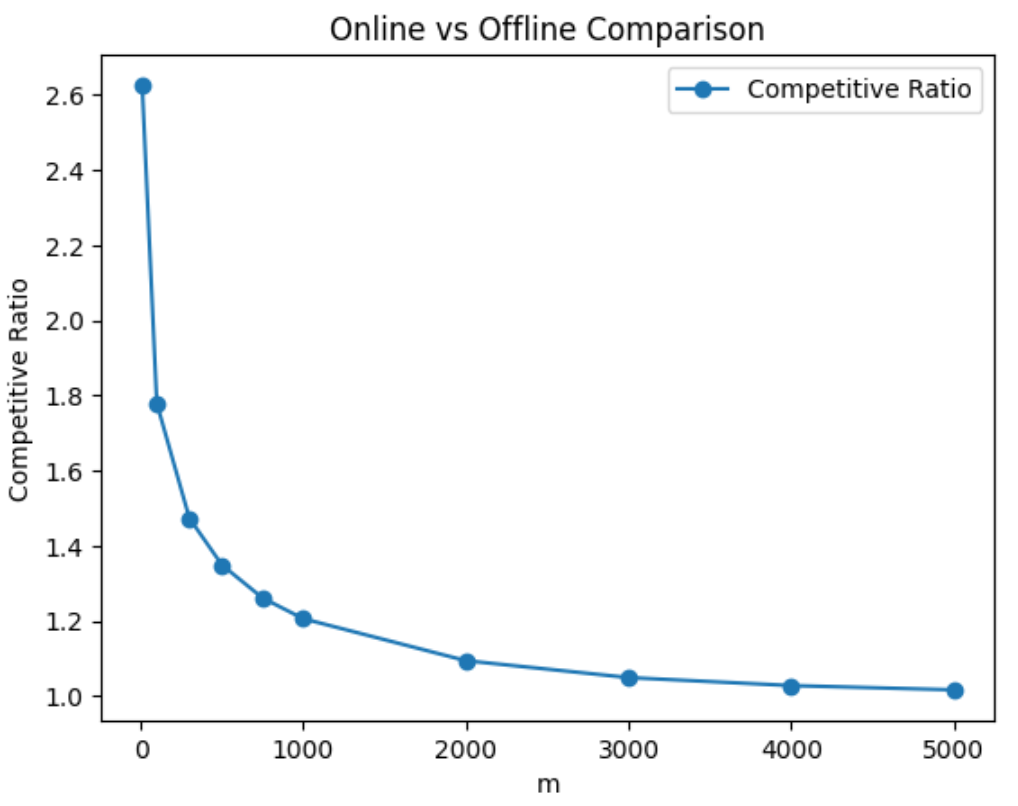} }}
    \caption{ Online Algorithm Experiments}
    \label{fig: online}
\end{figure}
\subsection{Online Algorithm}
In this part, we compare our online approach Algorithm \ref{alg:online} with the offline approach, i.e., Algorithm \ref{alg:modified-leak} with respect to the proxy loss $\mathbb{L}_m(\boldsymbol{c}, \boldsymbol{\epsilon})$. We demonstrate that as $m$ increases, the disparity between the online and offline loss reduces.

\noindent
\textbf{Implementation Details:} For our experiments, we consider that the privacy sensitivities $\boldsymbol{c}$ are drawn from the uniform distribution $\mathrm{U}[1,5]$. In Fig. \ref{fig: online}, we present a comparison of the results from both approaches for different values of the number of data points $m$. Similarly, we also plot the ratio of the online loss and the offline loss (i.e., the \emph{competitive ratio}) as a function of the number of samples $m$. 

\noindent
\textbf{Observations:} We observe that as $m$ increases, there is a consistent decrease in the loss, which aligns with our theoretical findings. Additionally, as $m$ increases, the competitive ratio approaches 1.

\section{Conclusion}
We introduce a novel algorithm to design a mechanism that balances competing objectives: achieving a high-quality logistic regression model consistent with differential privacy guarantees while minimizing payments to data providers. Notably, our result in Theorem \ref{gen loss} extends to scenarios where individual data points require different weights in loss calculations. This weighting accommodates noisy measurements or varying costs associated with sample retrieval. Additionally, our model considers heterogeneous privacy guarantees, recognizing the diverse privacy needs of individuals. Finally, Theorem \ref{mech design} shows that the payment mechanism does not depend on the choice of the loss function. Therefore, designing a payment mechanism and minimizing the objective can be effectively decoupled and treated as separate problems. This observation, combined with Theorem \ref{gen loss} (which highlights the need for additional regularization terms), opens avenues for designing mechanisms tailored to more complex ML problems. In addition, we also propose an online algorithm to design payments and differential privacy guarantees for sellers arriving sequentially.

As a future research direction, our work could be extended to model data markets of greater complexity. For instance, replacing the differential privacy training in Eq. (\ref{Eq: diff privacy eqn}) with a variant of DP-SGD, which handles heterogeneous differential privacy, could extend our framework to a broader class of machine learning models. Other future research could include relaxing the assumption that the distribution of privacy sensitivities is known beforehand for the online problem and instead learning it in an online fashion. Finally, additional promising avenues could explore correlations between data and privacy sensitivities or situations where data providers may also misrepresent their data.

\section{Data Dependent Payments}
\label{sec: data dependent pay}
In this section, we provide a computational method to solve a surrogate for the offline mechanism design objective Eq. \eqref{Eq: buyers obj2} in a nonasymptotic regime. 
\subsection{Objective Function}
To proceed with the algorithm formulation, as before, using Theorem \ref{gen loss}, the buyer's objective can be upper-bounded as
\begin{align}
&  \min_{\boldsymbol{w}, \boldsymbol{a}, \eta} \bigg[\sum_{i=1}^{m} a_{i} \log(1+e^{-y^i \cdot \boldsymbol{w}^T \boldsymbol{x}^i}) + \frac{2\boldsymbol{b}^T \boldsymbol{w}}{\eta} \nonumber \\ \label{objective}
 &\qquad+ \frac{\Lambda}{2}\|\boldsymbol{w}\|^2
+ \mu \|\boldsymbol{a}\|^2 + \sigma \frac{1}{\eta}   +\gamma  \eta \sum_{i=1}^{m}a_i\psi(c_i)\bigg],\\ 
&\mbox{s.t.}\ \ \ \boldsymbol{a} \geq 0,\ \eta \geq 0,\ \sum_{i=1}^{m} a_i = 1, \ a_i \leq \frac{k}{m},\ a_i \eta = \epsilon_i\ \forall i. \nonumber
\end{align}
Since it is mathematically intractable to represent the misclassification loss $\mathbb{E}[\mathbb{I}_{\{sign(\boldsymbol{w}^T \boldsymbol{x}) \neq y\}}]$, we consider Eq. \eqref{objective} as a surrogate for the buyer's objective. Therefore, we need to optimize Eq. \eqref{objective} jointly wrt $(\boldsymbol{w}, \boldsymbol{a}, \eta).$

Additionally, note that to make the objective function strictly convex, we have introduced a slight change in the objective by replacing $\|\boldsymbol{a}\|$ with $\|\boldsymbol{a}\|^2$. We further note that since $\|\boldsymbol{a}\|$ is a regularization term associated with the generalization error for the logistic loss, replacing $\|\boldsymbol{a}\|$ with $\|\boldsymbol{a}\|^2$ is aligned with our overall objective. The following theorem states that the new objective is strictly convex in $(\boldsymbol{w}, \boldsymbol{a})$ for sufficiently large $\Lambda$, and thus gradient descent converges exponentially fast to a global infimum.

\begin{theorem}
\label{convexify}
 Given a classification task, let  $D$ be a set of data points from $m$ users with $\|\boldsymbol{x}^i\| \leq 1, \forall i$. Then, the objective function as defined in Eq. (\ref{objective}) is strictly convex in $(\boldsymbol{w},\boldsymbol{a})$ if $2\Lambda \mu > m$ for any $\sigma, \eta \in \mathbb{R}_{+}$. Furthermore, $(\boldsymbol{w}^t,\boldsymbol{a}^t)_{t \in \mathbb{N}}$ be the sequence of iterates obtained by applying projected gradient descent on $f(\cdot)$ for a fixed $\eta$ on the set $S$, and let $f^*_{\eta}=\inf_{\boldsymbol{w},\boldsymbol{a}}f(\boldsymbol{w},\boldsymbol{z},\eta)$. Then, for $2\Lambda \mu > m$, there exists $0 < \alpha < 1$, such that
 \begin{equation*}
f(\boldsymbol{w}^t,\boldsymbol{a}^t,\eta) - f^*_{\eta}
\leq \alpha^t (f(\boldsymbol{w}^0,\boldsymbol{a}^0,\eta) - f^*_{\eta}).
 \end{equation*}
\hfill $\blacksquare$
\end{theorem} 

\smallskip
Note that the condition $2\Lambda \mu > m$, is a sufficient condition for strict convexity. Additionally, the convexity proof does not assume any structure on the dataset $D$ and therefore considers the worst-case possibility. However, in practice, smaller values of $\Lambda, \mu$, would be sufficient for making the objective strictly convex. In fact, we see in our experiments that the projected gradient descent on $(\boldsymbol{w},\boldsymbol{a})$ for a fixed $\eta$ converges to the same stationary point for different initializations. This suggests that the condition on $\mu, \Lambda$ in Theorem \ref{convexify} may be restrictive and it may be an interesting furture direction to relax this condition.



 \begin{algorithm}[H]\caption{An Iterative Algorithm to Optimize the Mechanism Design Objective}\label{alg:mechanism-base}
Set step-size $\alpha \in (0, 1],\ f_{\min} = \infty$. Discretize $[0,L]$ to any required precision and choose $\epsilon_{\text{avg}}$ from this discrete set.\\
    Sample $\|\boldsymbol{b}\|$ from $\Gamma(n,1)$ with its direction chosen uniformly at random.\\
{\bf For each} $\epsilon_{\text{avg}}$ {\bf do} \\
    Initialize $\boldsymbol{w}$ from $\mathbb{N}(0,1)$, $a_i = \frac{1}{m} \ \forall i$;\\
    {\bf While convergence do}\\
        $\boldsymbol{w} \leftarrow \boldsymbol{w} - \alpha \frac{d}{d\boldsymbol{w}}f(\boldsymbol{w},\boldsymbol{a},\epsilon_{\text{avg}}),$\\
        $\boldsymbol{a} \leftarrow \boldsymbol{a} - \alpha \frac{d}{d\boldsymbol{a}}f(\boldsymbol{w},\boldsymbol{a},\epsilon_{\text{avg}}),$\\
        $\boldsymbol{a} \leftarrow \mbox{\text{Proj}}_S(\boldsymbol{a})$ s.t. $S = \{\boldsymbol{a}: \sum_{i=1}^{m} a_i = 1\}$.\\
        {\bf end}\\
        {\bf If} $f_{\min} > f(\boldsymbol{w},\boldsymbol{a},\epsilon_{\text{avg}})$, {\bf then} \\
        { $\boldsymbol{w}_{\text{opt}} \gets \boldsymbol{w},$\ \ $\boldsymbol{a}_{\text{opt}} \gets \boldsymbol{a},$\ \
        $\epsilon_{\text{opt}} \gets \epsilon_{\text{avg}},$\ \ 
        $f_{\min} \gets f(\boldsymbol{w},\boldsymbol{a},\epsilon_{\text{avg}})$}.\\
   {\bf end}\\
{\bf end}
\end{algorithm}
\subsection{Algorithm}
We first optimize $f(\boldsymbol{w}, \boldsymbol{a}, \eta)$ with respect to $(\boldsymbol{w}, \boldsymbol{a})$ for a fixed $\eta$ over the constraint set using projected gradient descent, and then perform a line search over the scalar parameter $\eta$. To determine the range of $\eta$, note that $\eta = \sum_{i=1}^{m} \epsilon_i = m \epsilon_{avg}$, where $\epsilon_{avg} = \sum_{i} \epsilon_i / m$. Since the differential privacy guarantees $\boldsymbol{\epsilon}$ cannot be excessively high in practice, we restrict the range of $\boldsymbol{\epsilon}$ by taking $\epsilon_{avg} \in [0,L]$ for some $L \in \mathbb{R}_{+}.$ Thus, we can choose different values of $\eta$ by discretizing $[0,L]$ to any required precision and choosing $\epsilon_{avg}$ from it. The pseudocode is given in Algorithm \ref{alg:mechanism-base}. 

 Subsequently, the platform can pick an appropriate value of $\gamma$ by comparing different combinations of payment sum ($\sum_i t_i(D,\boldsymbol{c'})$) and misclassification loss ($\mathbb{E}[\mathbb{I}_{\{sign(\boldsymbol{w}^T \boldsymbol{x}) \neq y\}}]$) corresponding to each value of $\gamma$.

\section{Omitted Proofs}

\subsection{Proposition \ref{algoproof}}
\begin{proof}
Let $\boldsymbol{d}$ and $\boldsymbol{d'}$ be two vectors over $\mathbb{R}^n$ with norm at most $1$, and $y$, $y'$ being either $-1$ or $1$. Consider two different inputs given by $D := \{(\boldsymbol{x}^1,y^1),\ldots, (\boldsymbol{x}^{m-1},y^{m-1}), (\boldsymbol{d}, y)\}$ and $D' := \{(\boldsymbol{x}^1,y^1),\ldots, (\boldsymbol{x}^{m-1},y^{m-1}), (\boldsymbol{d'}, y')\}$. Since the function $\hat{\mathbb{L}}_m(D,\boldsymbol{w},\boldsymbol{a},\eta)$ is strictly convex in $\boldsymbol{w}$, for every $\boldsymbol{b'} = \frac{2\boldsymbol{b}}{\eta}$, there is a unique output $\hat{\boldsymbol{w}}$ for each input. Denote the values of $\boldsymbol{b'}$ for the first and second input such that the optimal solution is $\hat{\boldsymbol{w}}$ by $\boldsymbol{b'}_1$ and $\boldsymbol{b'}_2$, respectively, with the corresponding densities $h(\boldsymbol{b'}_1)$ and $h(\boldsymbol{b'}_2)$. By the optimality condition for convex functions the derivative of $\hat{\mathbb{L}}_m(D,\boldsymbol{w},\boldsymbol{a},\eta)$ must be 0 wrt $\boldsymbol{w}$ at $\hat{\boldsymbol{w}}$. Thus, we have
 \begin{equation}\nonumber
     \boldsymbol{b'}_1 - \frac{a_m \cdot \boldsymbol{d}y}{1+e^{y \hat{\boldsymbol{w}}^T \boldsymbol{d}}} = \boldsymbol{b'}_2 - \frac{a_m \cdot \boldsymbol{d'}y'}{1+e^{y' \hat{\boldsymbol{w}}^T \boldsymbol{d'}}}.
 \end{equation}
 Since $\frac{1}{1+e^{y \hat{\boldsymbol{w}}^T \boldsymbol{d}}} < 1$ and $\frac{1}{1+e^{y' \hat{\boldsymbol{w}}^T \boldsymbol{d'}}} < 1$, we have $\|\boldsymbol{b'}_1-\boldsymbol{b'}_2\| < 2 a_m$, which implies $\big| \|\boldsymbol{b'}_1\| - \|\boldsymbol{b'}_2\| \big| < 2 a_m$. Further, let $\mathbb{A}(D)$ be the output of our algorithm given dataset $D$ and $g(\boldsymbol{\hat{w}}|D)$ be the corresponding pdf. Therefore, for any two neighboring datasets $D$ and $D'$ using \cite{billingsley} we have
 \begin{equation}
 \label{Eq: bijection}
     \frac{g(\boldsymbol{\hat{w}}|D)}{g(\boldsymbol{\hat{w}}|D')} = \frac{h(\boldsymbol{b'}_1)}{h(\boldsymbol{b'}_2)} \cdot \frac{|det(J(\hat{\boldsymbol{w}}  \rightarrow   \boldsymbol{b'}_1|D))|^{-1}}{|det(J(\hat{\boldsymbol{w}}  \rightarrow \boldsymbol{b'}_2|D'))|^{-1}}.
 \end{equation}
 where $J(\hat{\boldsymbol{w}}  \rightarrow   \boldsymbol{b'}_1|D)$ is the Jacobian matrix of the mapping from $\boldsymbol{w}$ to $\boldsymbol{b}$.
 
 We first bound the ratio of the determinants. Using the optimality condition that the gradient at $\hat{\boldsymbol{w}}$ of $\hat{\mathbb{L}}_m(D,\boldsymbol{w},\boldsymbol{a},\eta)$ should be 0, we have
  \begin{equation}\nonumber
     \boldsymbol{b'}_1 =  \sum_{i=1}^{m} \frac{a_i \cdot \boldsymbol{x}^iy^i}{1+e^{y^i \hat{\boldsymbol{w}}^T \boldsymbol{x}^i}} - \lambda \hat{\boldsymbol{w}}
 \end{equation}
 Thus, taking the gradient of $\boldsymbol{b'}_1$ with respect to $\hat{\boldsymbol{w}}$, the Jacobian $J(\hat{\boldsymbol.{w}}  \rightarrow   \boldsymbol{b'}_1|D)$ matrix is given by
 \begin{equation}
 J(\hat{\boldsymbol{w}}  \rightarrow   \boldsymbol{b'}_1|D) = \sum_i \frac{-a_i e^{y^i \hat{\boldsymbol{w}}^T \boldsymbol{x}^i} \cdot \boldsymbol{x}^{i}(\boldsymbol{x}^{i})^T}{(1+e^{y^i \hat{\boldsymbol{w}}^T \boldsymbol{x}^{i}})^2} - \lambda I.
 \end{equation}

For simplicity, let us define two matrices $A$ and $E$ such that
 \begin{equation}
 A := -J(\hat{\boldsymbol{w}}  \rightarrow   \boldsymbol{b'}_1|D),
 \end{equation}
 \begin{align}
 E &:= J(\hat{\boldsymbol{w}}  \rightarrow   \boldsymbol{b'}_1|D) -J(\hat{\boldsymbol{w}}  \rightarrow   \boldsymbol{b'}_2|D') \nonumber \\
 E &=  \frac{-a_m e^{y^m \hat{\boldsymbol{w}}^T \boldsymbol{d'}^m} \cdot \boldsymbol{d'}(\boldsymbol{d'})^T}{(1+e^{y^m \hat{\boldsymbol{w}}^T \boldsymbol{d'}})^2} - \frac{-a_m e^{y^m \hat{\boldsymbol{w}}^T \boldsymbol{d}^m} \cdot \boldsymbol{d}(\boldsymbol{d})^T}{(1+e^{y^m \hat{\boldsymbol{w}}^T \boldsymbol{d}})^2} .
 \end{align}
 Now, we have
 \begin{equation}
  \frac{|det(J(\hat{\boldsymbol{w}}  \rightarrow   \boldsymbol{b'}_1|D))|^{-1}}{|det(J(\hat{\boldsymbol{w}}  \rightarrow \boldsymbol{b'}_2|D'))|^{-1}} = \frac{|det(A+E)|}{|det(A)|}.
 \end{equation}
 
 Let $\lambda_1(M)$ and $\lambda_2(M)$ denote the first and second largest eigenvalues of a matrix $M$. Since, $E$ is of rank 2, from Lemma 10 of \cite{kamalika}, we have
 \begin{align}\nonumber
  \frac{|det(A+E)|}{|det(A)|} &= | 1+ \lambda_1(A^{-1}E) + \lambda_2(A^{-1}E)\cr &\qquad+\lambda_1(A^{-1}E)\lambda_2(A^{-1}E) |. 
 \end{align}
Since we consider logistic loss which is $\Lambda$-strictly convex, any eigenvalue of A is atleast $\Lambda$. Thus $|\lambda_j(A^{-1}E)| \leq \frac{1}{\Lambda}|\lambda_j(E)|$. Now, applying the triangle inequality to the trace norm, we have
 \begin{equation}
 |\lambda_1(E)| + |\lambda_2(E)| \leq 2 ||d||^2 a_m \frac{e^{y^m \hat{\boldsymbol{w}}^T \boldsymbol{d}^m}}{(1+e^{y^m \hat{\boldsymbol{w}}^T \boldsymbol{d}^m})^2} \leq 2 a_m.
 \end{equation}
 Using Inequality of Arithmetic and Geometric means $\lambda_1(E)\lambda_2(E) \leq a_m^2$. Thus
 \begin{equation}
 \label{eq: jacoian ratio}
 \frac{|det(A+E)|}{|det(A)|} \leq \Big(1+\frac{a_m}{\Lambda}\Big)^2.
 \end{equation}
 Therefore, using Eq. (\ref{Eq: bijection}) and Eq. (\ref{eq: jacoian ratio}) we have
 \begin{align}
   \frac{g(\boldsymbol{\hat{w}}|D)}{g(\boldsymbol{\hat{w}}|D')} &\leq e^{\eta  (\|\boldsymbol{b'}_1\|-\|\boldsymbol{b'}_2\|) / 2} \cdot \frac{|det(A+E)|}{|det(A)|}  \nonumber \\
   &\leq \exp\Big(a_m \eta  + 2 \log\big(1 + \frac{k}{m \Lambda}\big)\Big) \nonumber \\
   &\leq \exp(\epsilon_m +  2 \log\big(1 + \frac{k}{m \Lambda}\big) \Big) \leq \exp(\epsilon_m + \Delta),
 \end{align}
 where the last inequality holds because $(\boldsymbol{a},\eta) \in \mathbb{F}$.
 \end{proof}
\subsection{Theorem \ref{gen loss}}
\begin{proof} For any $\boldsymbol{w} \in \mathbb{R}^n$ such that $||\boldsymbol{w}|| \leq \beta$ for some $\beta > 0$ and any sample set $S$=$\big\{(\boldsymbol{x}^1, y^1),\ldots,(\boldsymbol{x}^m, y^m)\big\}$ of $m$ iid datapoints sampled from the distribution $\mathcal{D}$
 we define the empirical loss function as
\begin{equation*}
    \hat{\mathcal{L}}_{S}[\boldsymbol{w}] := \sum_{i=1}^{m} a_i \log(1+e^{-y^i \cdot \boldsymbol{w}^T\boldsymbol{x}^i}) + \boldsymbol{b'}^T \boldsymbol{w}, 
\end{equation*}
where $\|\boldsymbol{b'}\| \sim \Gamma(n,\frac{2}{\eta})$, and the direction of $\boldsymbol{b'}$ is chosen uniformly at random. The true loss function is given by
\begin{equation*}
      \mathbb{E}_{(\boldsymbol{x},y)\sim \mathcal{D}}[\log(1+e^{-y\cdot \boldsymbol{w}^T\boldsymbol{x}})] = \mathcal{L}_m[\boldsymbol{w}].
\end{equation*}
Since $\sum_{i}^{} a_i = 1$ and $\mathbb{E}[\boldsymbol{b}] = 0$, we have
\begin{align}
\mathbb{E}\big[\mathbb{\hat{L}}_{S}[\boldsymbol{w}]\big] &= \sum_{i}^{} a_i \mathbb{E}[\log(1+e^{-y\cdot \boldsymbol{w}^T\boldsymbol{x}})] + \mathbb{E}[\boldsymbol{b'}^T\boldsymbol{w}] \nonumber \\
&= \mathbb{E}[\log(1+e^{-y\cdot \boldsymbol{w}^T\boldsymbol{x}})] = \mathcal{L}_m[\boldsymbol{w}].
\end{align}
Therefore, to upper bound the true loss we define, $\phi(S) = \sup_{\boldsymbol{w} \in \mathbb{R}^n}(\mathcal{L}_m[\boldsymbol{w}]-\mathbb{\hat{L}}_S[\boldsymbol{w}])$. Next, we will bound $\phi(S)$ using McDiarmids inequality. Now, for McDiarmids inequality to hold, we need to bound $\big|\boldsymbol{b'}^T \boldsymbol{w}\big|$. Therefore, we consider the event where $\big|\boldsymbol{b'}^T \boldsymbol{w}\big| < r$. This should be true for all $\boldsymbol{w}$ such that $\|\boldsymbol{w}\| \leq \beta$. This is possible only when $\|\boldsymbol{b'}\| < r/\beta$. Consider that this event happens with probability $1-\delta'$. From the CDF of $\Gamma(n,\frac{2}{\eta})$, we get 
 \begin{equation}
     \sum_{i=0}^{n-1} \frac{(\frac{\eta r}{2 \beta})^i}{i!} e^{-\frac{\eta r}{2 \beta}} = \delta'.
 \end{equation}
 Let $\eta r/\beta = t$ and $v(t) := \sum_{i=0}^{n-1} \frac{(t/2)^i}{i!} e^{-\frac{t}{2}}$. Also, it is known that $v(t)$ is a monotonously decreasing function. Therefore, its inverse exists, and we have
 \begin{equation}\nonumber
     r = \frac{\beta v^{-1}(\delta')}{\eta}.
 \end{equation}
Thus, by McDiarmid's inequality, the following holds with probability at least $(1-\delta-\delta')$
\begin{align}
   &\mathbb{P}(|\phi(S)-\mathbb{E}_S[\phi(S)] |> t)  \nonumber \\
    &\leq \exp\Bigg(\frac{-2t^2}{\sum_{}^{} a_i^2 \log^2(1+e^{\beta}) + (\frac{2\beta v^{-1}(\delta')}{\eta})^2}\Bigg).
\end{align}
Thus, with probability at least $(1-\delta-\delta')$,
\begin{equation}
    \phi(S) \leq \mathbb{E}_S[\phi(S)] + \sqrt{\frac{\ln \frac{1}{\delta}\Big(\sum a_{i}^2\log^2(1+e^{\beta}) + \big(\frac{2\beta v^{-1}(\delta')}{\eta})^2\Big)}{2}}.
\end{equation}
Moreover, we can write
\begin{align}\nonumber
    \mathbb{E}_S[\phi(S)] &= \mathbb{E}_S\big[\sup_{\|w\| \leq \beta} (\mathcal{L}_m[\boldsymbol{w}] - \mathbb{\hat{L}}_S(\boldsymbol{w}))\big] \cr
    &= \mathbb{E}_S\big[\sup_{\|w\| \leq \beta} \mathbb{E}_{S'}[\hat{\mathcal{L}}_{S'}(\boldsymbol{w}) - \mathbb{\hat{L}}_S(\boldsymbol{w})]\big] \cr
    &\leq \mathbb{E}_{S,S'}\big[\sup_{\|w\| \leq \beta} [\mathbb{\hat{L}}_{S'}(\boldsymbol{w}) - \mathbb{\hat{L}}_S(\boldsymbol{w})]\big] .
\end{align}
where the last inequality holds using Jensen's inequality. Next, we define iid Rademacher random variables $\sigma_i$. Now, since $S$ and $S'$ are drawn independently from $\mathcal{D}$ by symmetric, we can write the result in terms of $\sigma_i$ as follows
\begin{align}
    \mathbb{E}_S[\phi(S)] &\leq \mathbb{E}_{S,S',\sigma_i}\big[\sup_{\|w\| \leq \beta} [\sum_{i=1}^{m} a_i \sigma_i \log(1+e^{-y^i \cdot \boldsymbol{w}^T\boldsymbol{x}^i}) \nonumber \\
    &- \sum_{i=1}^{m} a_i \sigma_i \log(1+e^{-y'^i \cdot \boldsymbol{w}^T\boldsymbol{x'}^i})]\big] \nonumber \\
    &\leq \mathbb{E}_{S,\sigma_i} \bigg[\sup_{\|w\| \leq \beta}\Big[\sum_{i=1}^{m} a_i \sigma_i \log(1+e^{-y^i \cdot \boldsymbol{w}^T\boldsymbol{x}^i})\Big] \bigg] \nonumber \\
    &+ \mathbb{E}_{S',\sigma_i} \bigg[\sup_{\|w\| \leq \beta}\Big[\sum_{i=1}^{m} -a_i \sigma_i \log(1+e^{-y'^i \cdot \boldsymbol{w}^T\boldsymbol{x'}^i})\Big] \bigg]\cr
    &= 2 R_m(\boldsymbol{w}), \nonumber
\end{align}
where the last inequality holds by the sub additivity of the supremum function. Moreover, $R_m(\boldsymbol{w})$ is given by 
\begin{equation*}
    R_m(\boldsymbol{w}) := \mathbb{E}_{\sigma,S}[\sup_{\|w\| \leq \beta} \sum_{i=1}^{m} a_i \sigma_i \log(1+e^{-y^i \cdot \boldsymbol{w}^T\boldsymbol{x}^i})],
\end{equation*}
Now, we can again use McDiarmid's inequality to get
\begin{align}\nonumber
    &R_m(\boldsymbol{w}) \nonumber \\
    &\leq \hat{R}_S(\boldsymbol{w}) +\sqrt{\frac{\ln \frac{1}{\delta}\Big(\sum_i a_{i}^2\log^2(1+e^{\beta}) + \big(\frac{2\beta v^{-1}(\delta')}{\eta})^2\Big)}{2}}\cr 
    &\!\leq\! \hat{R}_S(\boldsymbol{w})\!+\! \sqrt{\frac{\ln \frac{1}{\delta}}{2}} \Bigg(\sqrt{\sum a_{i}^2\log^2(1+e^{\beta})} + \frac{2\beta v^{-1}(\delta')}{\eta}\Bigg). \nonumber
\end{align}
where $\hat{R}_S(\boldsymbol{w}) := \mathbb{E}_\sigma[\sup_{\boldsymbol{w} \in R^n} \sum_{i}^{} a_i \sigma_i \log(1+e^{-y^i \boldsymbol{w}^T \boldsymbol{x}^i}) + \boldsymbol{b'}^T \boldsymbol{w}]$
Finally, we bound $\hat{R}_S(\boldsymbol{w})$ as
\begin{align}\nonumber
\hat{R}_S(\boldsymbol{w}) & \leq \frac{1}{\ln2} E_\sigma [\sup_{\boldsymbol{w}}  \sum_{i}^{} a_i \sigma_i (-y^i \boldsymbol{w}^T \boldsymbol{x}^i)] + \frac{\beta v^{-1}(\delta')}{\eta}\nonumber \\
    &\leq \frac{\|\boldsymbol{w}\|}{\ln2} \mathbb{E}_\sigma [ \sum_{i}^{} a_i \sigma_i \boldsymbol{x}^i] + \frac{\beta v^{-1}(\delta')}{\eta} \nonumber 
    \end{align}
\begin{align}
    &\leq \frac{\beta}{\ln2} \sqrt{\sum_{i}^{} a_i^2} + \frac{\beta v^{-1}(\delta')}{\eta}, \label{rademacher}
\end{align}
where the first inequality holds by the Lipschitz property, and the last inequality uses $\|\boldsymbol{x}\| \leq 1$ and $\|\boldsymbol{w}\| \leq \beta$. Putting it together, we have
\begin{align}
&\mathbb{E}[\mathbb{I}_{\{sign(\boldsymbol{w}^T \boldsymbol{x}) \neq y\}}] \leq \mathbb{E}_{(\boldsymbol{x},y)\sim \mathcal{D}}[\log(1+e^{-y\cdot \boldsymbol{w}^T\boldsymbol{x}})] \nonumber \\
     &\leq \sum_{i=1}^{m} a_{i} \log(1+e^{-y^i  \boldsymbol{w}^T\boldsymbol{x}^i)}) + \boldsymbol{b'}^T \boldsymbol{w}  \nonumber \\
     &+ \Bigg[\Big(\frac{3  \ln \frac{1}{\delta}}{\sqrt{2}}\Big) \log(1+e^{\beta}) + \frac{\beta}{\ln2}\Bigg] \sqrt{\sum_{i}^{} a_i^2 } \cr
    &\qquad+ \Big(\frac{6  \ln \frac{1}{\delta}}{\sqrt{2}} + 1\Big) \Big(\frac{2\beta v^{-1}(\delta')}{\eta}\Big) + \frac{\Lambda}{2} ||\boldsymbol{w}||^2.
\end{align}
We also add the non negative term $\Lambda/2 ||\boldsymbol{w}||^2$ in the upper bound to get the desired result.
Thus, it is enough to define 
$\mu(\delta,\beta) = \Big(\frac{3  \ln \frac{1}{\delta}}{\sqrt{2}}\Big) \log(1+e^{\beta}) + \frac{\beta}{\ln2}$ and $\sigma(\delta,\delta',\beta) = \Big(\frac{6  \ln \frac{1}{\delta}}{\sqrt{2}} + 1\Big) \Big(2\beta v^{-1}(\delta')\Big)$.
\end{proof}
\subsection{Lemma \ref{mech design}}
\begin{proof} The proof follows similar steps as those in \cite{Asu}. Let $h_{i}(c)$ = $\mathbb{E}_{\boldsymbol{c}_{-i}}[\mathcal{L}_m(D,\boldsymbol{c};\boldsymbol{\epsilon},\boldsymbol{\theta})]$,
where $c$ is the argument corresponding to the privacy of agent $i$. Similarly, let $t_{i}(c)$ = $\mathbb{B}_{\boldsymbol{c}_{-i}}[t_{i}(c,\boldsymbol{c}_{-i})]$ and $\epsilon_{i}(c)$ = $\mathbb{E}_{\boldsymbol{c}_{-i}}[\epsilon_{i}(c,\boldsymbol{c}_{-i})]$.
Using the IC constraint, we have 
\begin{equation}\nonumber
   c_{i}\cdot \epsilon_{i}(c_{i}) - t_{i}(c_{i}) \leq  c_{i}\cdot \epsilon_{i}(c'_{i}) - t_{i}(c'_{i}).
\end{equation}
From the IC constraint, the function $c_{i}\cdot \epsilon_{i}(c) - t_{i}(c)$ has a minima at $c=c_{i}$. Thus, by equating the derivative to 0 and substituting $c=c_i$, we get
\begin{equation}
    c_{i}\cdot \Big(\frac{d \epsilon_{i}(c)}{dc}\Big)_{c=c_i} = t'_{i}(c_i).
\end{equation}
Solving for $c_i$ from this equation we get,
\begin{equation}
\label{IC-IR}
    t_{i}(c_{i}) = t_{i}(0) + c_{i}\epsilon_{i}(c_{i}) - \int_{0}^{c_{i}}\epsilon_{i}(z) dz.
\end{equation}
If an individual does not participate in estimating the parameter by not giving their data, then their loss function will be 0.
Thus, using the IR constraint, for all $c_{i}$ we have
\begin{equation*}\nonumber
    t_{i}(0) \geq \int_{0}^{c_{i}}\epsilon_{i}(z) dz.
\end{equation*}
Because $\epsilon_{i}(c_{i}) \geq 0$, it implies
\begin{equation*}
    t_{i}(0) \geq \int_{0}^{\infty}\epsilon_{i}(z) dz
\end{equation*}
Plugging this relation into Eq (\ref{IC-IR}), we get
\begin{equation*}
    t_{i}(c_{i}) \geq c_{i}\epsilon_{i}(c_{i}) + \int_{c_{i}}^{}\epsilon_{i}(z) dz.
\end{equation*}
Thus, for given $\boldsymbol{c}$, the payments are calculated to be $c_{i}\epsilon_{i}(c_{i}) + \int_{c_{i}}^{}\epsilon_{i}(z) dz$. Also, note that the minimum cost required for $c_i = \infty$ would be 0. Therefore, this can also be written as $-\int_{c_i}{} z \frac{d }{dz} \epsilon_{i}(z) dz$. Now, we can compute $\mathbb{E}_{c_{i}}[t_{i}(c_{i})]$ as 
\begin{align}\nonumber
    \mathbb{E}_{c_{i}}[t_{i}(c_{i})]&= \mathbb{E}_{c_{i}}[c_{i}\epsilon_{i}(c_{i})] + \mathbb{E}_{c_{i}}[\int_{c_{i}}^{}\epsilon_{i}(z) dz]   \cr
    &= \int_{z_{-i}}^{} \int_{z_{i}}^{} \Big(z_{i} \epsilon_{i}(z_{i},z_{-i}) \nonumber \\
    &+
    \int_{y_{i}=z_{i}}^{} \epsilon_{i}(y,z_{-i}) dy_{i}\Big) f_{i}(z_{i})dz_{i}f_{-i}(z_{-i})dz_{-i}. \nonumber
\end{align} 
By changing the order of integrals, we have
\begin{equation}\nonumber
    \mathbb{E}_{c_{i}}[t_{i}(c_{i})] = \mathbb{E}_{\boldsymbol{c}}[\Psi_{i}(c_{i})\epsilon_{i}(\boldsymbol{c})],
\end{equation}
where $\Psi_{i}(c_{i}) = c_{i} + \frac{F_{i}(c_{i})}{f_{i}(c_{i})}$. 
which completes the proof. 


\end{proof}

  \subsection*{Auxiliary Lemmas for the Proof of Theorem \ref{thm: asymptotic}}
    We first state two important lemmas, which are crucial in the proof of Theorem \ref{thm: asymptotic}
  \begin{lemma}
      Let $f_{\Psi}(.)$ be the probability density function corresponding to the random variable $\psi(c)$. Then under Assumption \ref{assumption: pdf} and \ref{assumption: virtual cost}, $f_{\Psi}(.)$ exists and  $\exists \ c_1 > 0$ such that $0 < f_{\Psi}(c) < \infty \ $for$ \ c \in [0,c_1].$
  \end{lemma}
  \begin{proof}
     \begin{align}
      f_{\Psi}(x) &= \frac{d}{dx}\mathbb{P}(\psi(x) < x) \nonumber \\
      &= \frac{f_C(\psi^{-1}(x))}{\psi'(\psi^{-1}(x)}
  \end{align}
  Now, since under Assumption \ref{assumption: virtual cost}, $\psi$ is a strictly increasing function, therefore its inverse exists and is strictly increasing. Therefore, $f_{\Psi}(x)$ exists. Furthermore, using Assumption \ref{assumption: pdf}, we show that $\exists \ c_1 > 0$ such that $f_{\Psi}(c) > 0 \ \forall \ c \in [0,c_1].$
  \end{proof}
   \begin{lemma}
 \label{thm: offline perf}
For every $\mu,\sigma,\gamma > 0$ and $0 < r' < 1/4$,  if $\boldsymbol{\hat{\epsilon}} = \arg \min_{\boldsymbol{\epsilon}} \mathbb{L}_m(\boldsymbol{c},{\boldsymbol{\epsilon}})$ then for a fixed constant $E_2$,
  \begin{equation}
   \mathbb{L}_m(\boldsymbol{c},\boldsymbol{\hat{\epsilon}})\leq  E_2 \frac{\sigma^{1/4}\mu^{1/2}\gamma^{1/4}}{\sqrt{3} m^{1/4}} + \mathcal{O}(m^{-1/4 - r'})
 \end{equation}
 with probability at least $1-\bar{\delta}$ where $\bar{\delta} = \mathcal{O}\big(\exp(-m^{1/2-2r'})\big)$
 \end{lemma}
\begin{proof}
Let $\psi(c)$ be distributed according to the pdf $f_{\Psi}(c).$ Using Taylor series expansion around $0$, we can write
 \begin{align}
 \label{Eq: square expectaion}
&g(\lambda):=\mathbb{E}\big[\big((\lambda-\gamma\psi(\boldsymbol{c}))^{+}\big)^2\big] = \int_{0}^{\lambda/\gamma} \gamma^2(\lambda/\gamma-x)^2 f_{\Psi}(x) dx \nonumber \\
     &= g(0) +  g'(0)\lambda + \frac{ g''(0)}{2}\lambda^2 + \frac{g'''(l)}{6}\lambda^3 = \frac{f_{\Psi}(l)}{3\gamma}\lambda^3, 
 \end{align}
 where the last equality holds because a simple calculation shows that $g(0)=g'(0)=g''(0)=0$ and $g'''(l)=2f_{\Psi}(l)/\gamma$, where $l \in [0,\lambda/\gamma].$ Similarly,
 \begin{align}
 \label{Eq: sum expectation}
w(\lambda)&:=\mathbb{E}\big[\big((\lambda-\gamma\psi(\boldsymbol{c}))^{+}\big)\big]  = \int_{0}^{\lambda/\gamma} \gamma(\lambda/\gamma-x) f_{\Psi}(x) dx \nonumber \\
     &= w(0)+w'(0)\lambda + \frac{w''(l')}{2}\lambda^2  = \frac{ f_{\Psi}(l')}{2\gamma} \lambda^2,  
 \end{align}
 for some $l' \in [0,\lambda/\gamma]$.
 
 Next, using Eq. \eqref{Eq:lambda}, we know that
 \begin{align}\nonumber
         &\lambda - \frac{\sigma}{\mu^2} \|(\lambda-\gamma\psi(\boldsymbol{c}))^{+}\|^2 - \frac{\|(\lambda-\gamma\psi(\boldsymbol{c}))^{+}\|^2}{\sum_i (\lambda-\gamma\psi(c_i))^{+}} = 0.      
 \end{align}
Since solving the above equation 
in a closed-form is difficult, instead we form a surrogate equation in which $\sum_i (\lambda-\gamma\psi(c_i))^{+}$ and $\|(\lambda-\gamma\psi(\boldsymbol{c}))^{+}\|^2$ are replaced by their expected values. Therefore, the surrogate equation is given by
\begin{align} \label{Eq: lambda-surr}
            &\lambda - \frac{\sigma m}{\mu^2} \frac{\lambda^3 f_{\Psi}(l)}{3\gamma} - \frac{2\lambda^3 f_{\Psi}(l)}{3\lambda^2 f_{\Psi}(l')} = 0.
\end{align}
Let $\bar{\lambda}\neq 0$ be the nontrivial solution to the surrogate equation \eqref{Eq: lambda-surr}, which gives us an approximate solution to Eq. \eqref{Eq:lambda}. Note that we can also rewrite Eq. \eqref{Eq: lambda-surr} as 
 \begin{equation}
 \label{Eq: fixed pt}
     \lambda = \frac{1}{\sqrt{m}}\sqrt{\frac{3\gamma\mu^2}{\sigma f_{\Psi}(l)}\Big(1-\frac{2f_{\Psi}(l)}{3f_{\Psi}(l')}\Big)} := \frac{G}{\sqrt{m}}.
 \end{equation}

We first show that $\Big(1-\frac{2f_{\Psi}(l)}{3f_{\Psi}(l')}\Big) > 0$ for $\lambda > 0.$ Otherwise, using Eq. (\ref{Eq: sum expectation}) and Eq. (\ref{Eq: square expectaion}), we have
\begin{align}
    &\frac{2f_{\Psi}(l)}{3f_{\Psi}(l')} = \frac{\int_{0}^{\lambda/\gamma}(\lambda-\gamma x )^2 f_{\Psi}(x) dx}{\int_{0}^{\lambda/\gamma}\lambda(\lambda-\gamma x ) f_{\Psi}(x) dx} \geq 1 \nonumber \\
   \implies &\int_{0}^{\lambda/\gamma} (\lambda^2-2\lambda\gamma x + \gamma^2 x^2-\lambda^2 + \lambda \gamma x) f_{\Psi}(x) dx \geq  0 \nonumber \\
   \implies &\int_{0}^{\lambda/\gamma} \gamma x(\gamma x-\lambda) f_{\Psi}(x) dx \geq 0, 
\end{align}
which is a contradiction.

Next, we show that for sufficiently large $m$, the surrogate equation \eqref{Eq: fixed pt} has a solution. To show that, consider the function $e(\lambda) = \lambda - \frac{1}{\sqrt{m}}\sqrt{\frac{3\gamma\mu^2}{\sigma f_{\Psi}(l)}\Big(1-\frac{2f_{\Psi}(l)}{3f_{\Psi}(l')}\Big)}$. For $\lambda = 0$, $f_{\Psi}(l) = f_{\Psi}(l') = f_{\Psi}(0).$ Therefore, $e(0) < 0.$ Since $\exists \ c_1 > 0$ such that $f_{\Psi}(c) > 0 \ \forall \ c \in [0,c_1]$, we can pick $\lambda = \lambda_1$ such that $f_{\Psi}(c) > 0,$ for every $c \in [0,\lambda_1/\gamma].$ Therefore, for sufficiently large $m$, $\lambda_1 > \frac{1}{\sqrt{m}}\sqrt{\frac{3\gamma\mu^2}{\sigma f_{\Psi}(l)}} > \frac{1}{\sqrt{m}}\sqrt{\frac{3\gamma\mu^2}{\sigma f_{\Psi}(l)}\Big(1-\frac{2f_{\Psi}(l)}{3f_{\Psi}(l')}\Big)}.$ Hence, using the Intermediate Value theorem, there exists a solution to Eq. \eqref{Eq: fixed pt} for a sufficiently large $m$.
 
 Since we have now established that the surrogate equation has a nontrivial solution, we will use Eq. (\ref{Eq: fixed pt}) to get an upper bound for the excess risk $\mathbb{L}_m(\boldsymbol{c},\bar{\boldsymbol{\epsilon}})$. To that end, using Bernstein's inequality with probability $(1-\delta_1)$, where $\delta_1 = 2\exp\big(\frac{-\frac{1}{2}m\alpha'^2}{var\big(((\lambda - \gamma\psi(c))^{+})^2\big) + \frac{1}{3}\alpha' \lambda^2}\big)$, the following holds
 \begin{equation}\nonumber
     \frac{1}{m} \sum_{i=1}^{m} \big((\lambda-\gamma\psi(c_i))^{+}\big)^2 \leq \mathbb{E}\big[\big((\lambda-\gamma\psi(c_i))^{+}\big)^2\big] + \alpha'.
 \end{equation}
 To calculate $\delta_1$, we proceed to calculate $var\big(((\lambda - \gamma\psi(c))^{+})^2\big)$. 
 \begin{align}
&h(\lambda):=\mathbb{E}\big[\big((\lambda-\gamma\psi(\boldsymbol{c}))^{+}\big)^4\big] = \int_{0}^{\lambda/\gamma} \gamma^4(\lambda/\gamma-x)^4 f_{\Psi}(x) dx \nonumber \\
     &= h(0) +  \sum_{i=1}^{4} \left. \frac{\partial^i h(x)}{\partial x^i}\right|_{x=0}\frac{\lambda^i}{i!} + \left. \frac{\partial^5 h(x)}{\partial x^5}\right|_{x=l''}\frac{\lambda^5}{120} = \frac{f_{\Psi}(l'')}{5\gamma}\lambda^5, \nonumber
 \end{align}
 for some $l'' \in [0,\lambda/\gamma]$. Therefore, we have $$var\big(((\lambda - \gamma\psi(c))^{+})^2\big) = \frac{f_{\Psi}(l'')}{5\gamma}\lambda^5 - \frac{f^2_p(l)}{9\gamma^2}\lambda^6.$$
 
 Now, choosing $\alpha' = \frac{G_1 m^{\kappa'}}{m^{7/4}}$ for some $G_1 > 0$ and for $0 < \kappa' < 1/4$, we get
 \begin{align}
     \delta_1 &:= 2\exp \bigg(\frac{-G_1^2 m^{2\kappa'}}{\frac{2f_{\Psi}(l'')G^5}{5\gamma } - \frac{2f^2_p(l) G^6}{9\gamma^2 m^{1/2}} + \frac{2G^2 G_1}{3 m^{1/4-\kappa'}}}\bigg) \nonumber \\
     &= \mathcal{O}\big(\exp(-m^{2\kappa'})\big) = \mathcal{O}\big(\exp(-m^{1/2-2r'})\big),
 \end{align}
 where the last equality follows by choosing $\kappa = 1/4-r'$ for some $0<r'<1/4.$
Note that $G, f_{\Psi}(l'), f_{\Psi}(l)$ intrinsically depend on $m$ through their dependence on $\lambda.$ However, since $0 < f_{\Psi}(c) < \infty$ for $c \in [0,c_1]$, we can choose a sufficiently large $M'$ such that for $m > M'$, $f_{\Psi}(l), f_{\Psi}(l') \in [r_1,r_2]$ for some constants $r_1, r_2 > 0.$ Similarly, we can find constants $r_3, r_4 > 0$ such that $G \in [r_3,r_4].$ Thus, with a probability $(1-\delta_1),$ we have
  \begin{equation}
  \label{eq: square bound}
    \frac{1}{m} \sum_{i=1}^{m} \big((\lambda-\gamma\psi(c_i))^{+}\big)^2 \leq \mathbb{E}\big[\big((\lambda-\gamma\psi(c_i))^{+}\big)^2\big] + \frac{G_1}{m^{3/2+r'}}.
 \end{equation}
 Similarly, with probability at least $(1-\delta_2)$, we have
   \begin{equation}
   \label{eq: sum bound}
   \frac{1}{m} \sum_{i=1}^{m} (\lambda-\gamma\psi(c_i))^{+} \leq \mathbb{E}\big[(\lambda-\gamma\psi(c_i))^{+}\big] + \frac{G_2}{m^{1+r'}},
 \end{equation}
 where $\delta_2$ is given by
 \begin{equation}
     \delta_2 := \mathcal{O}\big(\exp(-m^{1/2-2r'})\big).
 \end{equation}
Next, we will upper-bound each of the terms in $\mathbb{L}_m(\boldsymbol{c},\boldsymbol{\bar{\epsilon}})$. Therefore, we have the following with probability atleast $1-\delta_1-\delta_2 := 1-\bar{\delta}$
\begin{align}
\label{Eq: bounding proc}
    &\frac{\mu \|\boldsymbol{\bar{\epsilon}}\|}{\sum_i \bar{\epsilon}_i} + \frac{\sigma}{\sum_i \bar{\epsilon}_i} + \gamma \sum_{i=1}^{m} \bar{\epsilon}_i \psi(c_i) \nonumber \\
    &= \frac{\mu ||(\bar{\lambda} - \gamma \psi(\boldsymbol{c}))^+||}{\sum_i (\bar{\lambda} - \gamma \psi(\boldsymbol{c}))} + \frac{\sigma ||(\bar{\lambda} - \gamma \psi(\boldsymbol{c}))^+||}{\mu} \nonumber \\
    &+ \frac{\gamma \mu \sum_i \psi(c_i) (\bar{\lambda} - \gamma \psi(\boldsymbol{c}))^+}{||(\bar{\lambda} - \gamma \psi(\boldsymbol{c}))^+|| \sum_i (\bar{\lambda} - \gamma \psi(\boldsymbol{c}))^+} \nonumber \\
    &\leq \frac{\mu \sqrt{\frac{f_{\Psi}(l)\bar{\lambda}^3}{3 \gamma}+\frac{G_1}{m^{3/2+r'}}}}{m^{1/2} \Big(\frac{f_{\Psi}(l') \bar{\lambda}^2}{2\gamma} - \frac{G_2}{m^{1+r'}}\Big)} + \frac{\sigma m^{1/2}\sqrt{\frac{f_{\Psi}(l)\bar{\lambda}^3}{3 \gamma}+\frac{G_1}{m^{3/2+r'}}}}{\mu} \nonumber \\
    &+ \frac{\mu G}{m \sqrt{\frac{f_{\Psi}(l)\bar{\lambda}^3}{3 \gamma}-\frac{G_1}{m^{3/2+r'}}}} \nonumber \\
    &= \frac{\mu \sqrt{\frac{f_{\Psi}(l)G^3}{3 m^{3/2}\gamma}+\frac{G_1}{m^{3/2+r'}}}}{m^{1/2} \Big(\frac{f_{\Psi}(l') G^2}{2m\gamma} - \frac{G_2}{m^{1+r'}}\Big)} + \frac{\sigma m^{1/2}\sqrt{\frac{f_{\Psi}(l)G^3}{3 m^{3/2}\gamma}+\frac{G_1}{m^{3/2+r'}}}}{\mu} \nonumber \\
    &+ \frac{\mu G}{m \sqrt{\frac{f_{\Psi}(l)G^3}{3 m^{3/2}\gamma}-\frac{G_1}{m^{3/2+r'}}}} \nonumber \\
    &= E_{2a} \frac{\sigma^{1/4}\mu^{1/2}\gamma^{1/4}}{\sqrt{3} m^{1/4}} + \mathcal{O}(m^{-1/4 - r'}) \nonumber \\
    &\leq E_2 \frac{\sigma^{1/4}\mu^{1/2}\gamma^{1/4}}{\sqrt{3} m^{1/4}} + \mathcal{O}(m^{-1/4 - r'})
\end{align}
where
\begin{align}
\label{Eq: const}
&E_{2a}:= \frac{2f_{\Psi}^{3/4}(l)}{f_{\Psi}(l')\big(3-2f_{\Psi}(l)/f_{\Psi}(l')\big)^{1/2}} \cr
&+ \frac{ \big(3-2f_{\Psi}(l)/f_{\Psi}(l')\big)^{3/2}}{f_{\Psi}^{1/4}(l)} 
    \cr 
    &+ \frac{ f_{\Psi}(l'')}{ f_{\Psi}(l')f_{\Psi}^{1/4}(l)\big(3-2f_{\Psi}(l)/f_{\Psi}(l')\big)^{1/2}} \nonumber 
    \end{align}
    \begin{align}
    &\leq \frac{2f_{\Psi}^{3/4}(r_2)}{f_{\Psi}(r_1)\big(3-2f_{\Psi}(r_2)/f_{\Psi}(r_2)\big)^{1/2}} \cr
    &+ \frac{ \big(3-2f_{\Psi}(r_1)/f_{\Psi}(r_1)\big)^{3/2}}{f_{\Psi}^{1/4}(r_1)} 
    \cr 
    &+ \frac{ f_{\Psi}(r_2)}{ f_{\Psi}(r_1)f_{\Psi}^{1/4}(r_1)\big(3-2f_{\Psi}(r_2)/f_{\Psi}(r_2)\big)^{1/2}}:= E_2.
\end{align}
The first equality is obtained using Eq. \ref{Eq: eta eqn}. The next inequality is obtained using Eq. (\ref{eq: sum bound}) and Eq. (\ref{eq: square bound}), and substituting the expressions for expectations using Eq. (\ref{Eq: sum expectation}) and Eq. (\ref{Eq: square expectaion}) and noting that $\bar{\lambda} - \gamma \psi(c_i) > 0$ when $\psi(c_i) < \bar{\lambda}/\gamma.$ Finally, we substitute $\bar{\lambda}$ to get the final bound.
\end{proof}

\subsection{Proof of Theorem \ref{thm: asymptotic}}
\begin{proof}
\begin{align}
 &\mathbb{E}_{(\boldsymbol{x},y)\sim\mathcal{D}}[\mathbb{I}_{\{sign(\boldsymbol{\hat{w}}^T \boldsymbol{x}) \neq y\}}] + \gamma \cdot \sum_{i=1}^{m}\psi(c_{i})\epsilon_{i}(\boldsymbol{c}) \nonumber \\
    &\leq \sum \hat{a}_i \log(1+e^{-y^i \boldsymbol{\hat{w}}^T \boldsymbol{x}^i}) + \frac{2\boldsymbol{b}^T\boldsymbol{\hat{w}}}{\hat{\eta}} \nonumber \\
    & + \mu (\delta,\beta) ||\boldsymbol{\hat{a}}|| + \sigma (\delta, \delta', \beta)  \frac{1}{\hat{\eta}} + \gamma \sum_{i=1} \hat{\epsilon}_i\psi(c_i)  \nonumber \\
    &\leq \sum \hat{a}_i \log(1+e^{-y^i \boldsymbol{\hat{w}}^T \boldsymbol{x}^i}) + \frac{2\boldsymbol{b}^T\boldsymbol{\hat{w}}}{\hat{\eta}} + E_2 \frac{\sigma^{1/4}\mu^{1/2}\gamma^{1/4}}{\sqrt{3} m^{1/4}} \nonumber \\
    &+ \mathcal{O}(m^{-1/4 + r'}) \nonumber\\
    &\leq \sum \hat{a}_i \log(1+e^{-y^i (\boldsymbol{w}^*\beta)^T \boldsymbol{x}^i}) + \frac{2\boldsymbol{b}^T\boldsymbol{w}^*\beta}{\hat{\eta}} \nonumber \\
    & + E_2 \frac{\sigma^{1/4}\mu^{1/2}\gamma^{1/4}}{\sqrt{3} m^{1/4}} + \mathcal{O}(m^{-1/4 - r'}) \nonumber \\
    &\leq \log(1+e^{-\rho\beta}) + E_2 \frac{\sigma^{1/4}\mu^{1/2}\gamma^{1/4}}{\sqrt{3} m^{1/4}} + \mathcal{O}(m^{-1/4 - r'}) \nonumber \\
    &\leq \log(1+e^{-\rho\beta}) + E'_2 \frac{\gamma^{1/4}}{\sqrt{3} m^{1/4-r}} + E''_2 \frac{\gamma^{1/4}}{\sqrt{3} m^{1/4}}\nonumber \\
    &+ \mathcal{O}(m^{-1/4+r-r'}) \nonumber \\
    &= \log(1+e^{-\rho\beta}) + \nu + \mathrm{o}(1),
\end{align}
where $E'_2 = E_2 \frac{3}{\sqrt{2}} \log(1+e^{\beta})$ and $E''_2 = E_2 \frac{\beta}{\ln 2}$
The first inequality follows from  Theorem \ref{gen loss} and holds with probability $(1-\delta-\delta')$. Using Lemma \ref{thm: offline perf}  we obtain the second inequality which holds with probability $(1-\delta-\delta'-\bar{\delta})$. Moreover, the third inequality holds because $\hat{\boldsymbol{w}}$ minimizes Eq. (\ref{Eq: diff privacy eqn}). Finally, from the assumption \ref{assumption: sep}, $\sum a_i = 1$ and $1/\eta = \mathcal{O}(m^{-1/4})$, we obtain the fourth inequality.

 Next, we choose $\delta = \exp(-E_3 \nu m^{r})$ for some $r > 0$ where $E_3 = \frac{\sqrt{2}}{\sqrt{3} E_2 \gamma^{1/4} \log(1+e^{\beta})} \approx \frac{\sqrt{2}}{\sqrt{3} E_2 \gamma^{1/4}\beta}$. Therefore, from  Thm. \ref{gen loss} $\mu = \nu + \frac{\beta}{\ln 2}$. Additionally, we choose $\delta'$ such that $\sigma = \mu^2$. Therefore, substituting in the expression for $\sigma$, we have $\delta' = \mathcal{O}(((\nu/\beta) m^r)^{n}e^{-\nu m^r/2\beta}) \sim \mathcal{O}(e^{-\nu m^{r}/2\beta})$. Substituting the value for $\mu$ and $\sigma$, we obtain the last inequality. Finally, we choose $r = 1/4, r' = 1/8$ to get the final result which holds with probability $1-3\delta'$ where $\delta' = \mathcal{O}(e^{-\nu m^{1/4}/2\beta}).$
\end{proof}

\subsection{Proof of Theorem \ref{thm: perf bounds}}
\begin{proof}
We will first upper bound Eq. (\ref{Eq: utility privacy tradeoff}) as before for $\tilde{\boldsymbol{\epsilon}}$ derived from Algorithm \ref{alg:online}. Calculations follow similar to Eq. (\ref{Eq: bounding proc}) with upper bounds obtained using Eq. (\ref{eq: sum bound}) and Eq. (\ref{eq: square bound}) and expressions for expectation taken using Eq. (\ref{Eq: sum expectation}), (\ref{Eq: square expectaion}).
\begin{align}
        &\mathbb{L}_m(\boldsymbol{c},\boldsymbol{\tilde{\epsilon}}) = \frac{\mu \|\boldsymbol{\tilde{\epsilon}}\|}{\sum_i \epsilon_i} + \frac{\sigma}{\sum_i \tilde{\epsilon_i}} + \gamma \sum_{i=1}^{m} \tilde{\epsilon_i} \psi(c_i)\nonumber \\
        &=\frac{\mu\|(\tilde{\lambda}-\psi(c_i))^{+}\|}{\sum_i (\tilde{\lambda}-\psi(c_i))^{+}} + \frac{\sigma f_{\Psi}^{3/2}(0) m^{3/2} \tilde{\lambda}^{7/2}}{2\sqrt{3}\gamma^{3/2}\mu\sum_i (\tilde{\lambda}-\gamma\psi(c_i))^{+}}\nonumber\\
    &+ \frac{2\sqrt{3}  \gamma^{3/2}\mu\sum_i \psi(c_i)(\tilde{\lambda}-\gamma \psi(c_i))^{+}}{f_{\Psi}^{3/2}(0) m^{3/2} \tilde{\lambda}^{7/2}} \nonumber \\
         &\leq \frac{2\mu\gamma^{1/2} f_{\Psi}^{1/2}(l)}{\sqrt{3} (\xi m)^{1/2} \tilde{\lambda}^{1/2}f_{\Psi}(l')} + \frac{\sigma  f_{\Psi}^{3/2}(0) m^{1/2} \tilde{\lambda}^{3/2}}{\sqrt{3}\gamma^{1/2}\xi \mu f_{\Psi}(l')} \nonumber\\
    &+ \frac{\sqrt{3} \mu \xi \gamma^{1/2} f_{\Psi}(l'')}{ f_{\Psi}^{3/2}(0) m^{1/2} \tilde{\lambda}^{1/2}} + \mathcal{O}( m^{-1/4-r'}) \nonumber \\
  &= \frac{\mu^{1/2}\sigma^{1/4} \gamma^{1/4}}{\sqrt{3} m^{1/4}}\Big[\frac{2 f_{\Psi}^{1/2}(l)f_{\Psi}^{1/4}(0) }{ \xi^{1/2} f_{\Psi}(l')} + \frac{  f_{\Psi}^{3/4}(0) }{ \xi f_{\Psi}(l')}\nonumber \\
    & + \frac{   3 \xi f_{\Psi}(l'')}{ f_{\Psi}^{5/4}(0) }\Big] + \mathcal{O}( m^{-1/4-r'})\nonumber \\
    &= E_4 \frac{\mu^{1/2}\sigma^{1/4} \gamma^{1/4}}{\sqrt{3} m^{1/4}} + \mathcal{O}( m^{-1/4-r'})
    \end{align}
    where $E_4 := \Big[\frac{2 f_{\Psi}^{1/2}(l)f_{\Psi}^{1/4}(0) }{ \xi^{1/2} f_{\Psi}(l')} + \frac{  f_{\Psi}^{3/4}(0) }{ \xi f_{\Psi}(l')} + \frac{   3 \xi f_{\Psi}(l'')}{ f_{\Psi}^{5/4}(0) }\Big]$
    In the last step we substitute $\lambda$ from Eq. (\ref{Eq: lambda-surr}). Thus, we see that $E_4$ is of the order $\mathcal{O}(1/\xi).$

    Therefore, as before,
    \begin{align}
        &\mathbb{E}_{(\boldsymbol{x},y)\sim\mathcal{D}}[\mathbb{I}_{\{sign(\boldsymbol{\hat{w}}^T \boldsymbol{x}) \neq y\}}] + \gamma \cdot \sum_{i=1}^{m}\psi(c_{i})\epsilon_{i}(\boldsymbol{c}) \nonumber \\
        &\leq
      \log(1+e^{-\rho\beta}) + E_4 \frac{\sigma^{1/4}\mu^{1/2}\gamma^{1/4}}{\sqrt{3} m^{1/4}} + \mathcal{O}(m^{-1/4 - r'}) \nonumber \\    
    &\leq \log(1+e^{-\rho\beta}) + E'_4 \frac{\gamma^{1/4}}{\sqrt{3} m^{1/4-r}} + E''_4 \frac{\gamma^{1/4}}{\sqrt{3} m^{1/4}}\nonumber \\
    &+ \mathcal{O}(m^{-1/4+r-r'}) \nonumber \\
    &= \log(1+e^{-\rho\beta}) + \nu + \mathrm{o}(1)
    \end{align}
where $E'_4 = E_4 \frac{3}{\sqrt{2}} \log(1+e^{\beta})$ and $E''_4 = E_4 \frac{\beta}{\ln 2}$

 Next, we choose $\delta = \exp(-E_5 \nu m^{r})$ for some $r > 0$ where $E_5 = \frac{\sqrt{2}}{\sqrt{3} E_4 \gamma^{1/4} \log(1+e^{\beta})} \approx \frac{\sqrt{2}}{\sqrt{3} E_4 \gamma^{1/4}\beta}$. Therefore, from  Thm. \ref{gen loss}, $\mu = \nu + \frac{\beta}{\ln 2}$. Additionally, we choose $\delta'$ such that $\sigma = \mu^2$. Therefore, substituting in the expression for $\sigma$, and noting that $E_5 = \mathcal{O}(\xi)$, we have $\delta' = \mathcal{O}(((\nu\xi/\beta) m^r)^{n}e^{-\nu\xi m^r/2\beta}) \sim \mathcal{O}(e^{-\nu\xi m^{r}/2\beta})$. Substituting the value for $\mu$ and $\sigma$, we obtain the last inequality. Finally, we choose $r = 1/4, r' = 1/8$ to get the final result which holds with probability $1-3\delta'$ where $\delta' = \mathcal{O}(e^{-\xi\nu m^{1/4}/2\beta}).$
   \end{proof}

   \subsection{Proof of Theorem \ref{convexify}}
\begin{proof} 
To prove that the objective function in Eq. \eqref{objective} is jointly convex in $(\boldsymbol{w}, \boldsymbol{a})$, we will calculate its Hessian. We denote the objective by $g(\boldsymbol{w},\boldsymbol{a})$. Therefore, $\nabla^2 g$ is given by
\begin{equation}
    \nabla^2 g = \begin{bmatrix} 
    \nabla^2 g_{\boldsymbol{w}, \boldsymbol{w}} & \nabla^2 g_{\boldsymbol{w}, \boldsymbol{a}}\\
    \nabla^2 g_{\boldsymbol{w}, \boldsymbol{a}} & \nabla^2 g_{\boldsymbol{a}, \boldsymbol{a}}
    \end{bmatrix}.
\end{equation}
Let vector $\boldsymbol{v} = [\boldsymbol{v_1} \ \boldsymbol{v_2}]$ be such that $\|\boldsymbol{v}\| = 1$. Therefore, for all such $\boldsymbol{v}$, we show that $\boldsymbol{v}^T \nabla^2 g \boldsymbol{v} > 0$. We have 
\begin{align}
\boldsymbol{v}^T \nabla^2 g \boldsymbol{v} &= \boldsymbol{v_1}^T \nabla^2 g_{\boldsymbol{w},\boldsymbol{w}} \boldsymbol{v_1} + \boldsymbol{v_2}^T \nabla^2 g_{\boldsymbol{a},\boldsymbol{a}} \boldsymbol{v_2} + 2 \boldsymbol{v_1}^T \nabla^2 g_{\boldsymbol{w},\boldsymbol{a}} \boldsymbol{v_2} \nonumber \\
&\geq \Lambda ||v_1||^2 + 2\mu ||v_2||^2 + \sum_i \frac{a_i e^{y^i \boldsymbol{w}^T \boldsymbol{x}^i}}{(1+ e^{y^i \boldsymbol{w}^T \boldsymbol{x}^i})^2} ((\boldsymbol{x}^i)^T \boldsymbol{v_1})^2\nonumber \\
&+ 2  \boldsymbol{v_1}^T \nabla^2 g_{\boldsymbol{w},\boldsymbol{a}} \boldsymbol{v_2} 
\end{align}
where
\begin{align}
    \nabla^2 g_{\boldsymbol{w},\boldsymbol{a}} = \begin{bmatrix}
        \frac{-y^1 e^{-y^1 \boldsymbol{w}^T\boldsymbol{x}^1}}{1 + e^{-y^1 \boldsymbol{w}^T\boldsymbol{x}^1}} (\boldsymbol{x}^1)^T  \\
      \ldots \\
            \frac{-y^m e^{-y^m \boldsymbol{w}^T\boldsymbol{x}^m}}{1 + e^{-y^m \boldsymbol{w}^T\boldsymbol{x}^m}} (\boldsymbol{x}^m)^T  \\
    \end{bmatrix}.
\end{align}
Therefore,
\begin{align}
  \boldsymbol{v}^T \nabla^2 g \boldsymbol{v} &\geq \Lambda ||v_1||^2 + 2\mu ||v_2||^2 \nonumber \\
  &- 2 \sum_i v_1(i) y^i \frac{ e^{-y^i\boldsymbol{w}^T\boldsymbol{x}^i}}{1 + e^{-y^i \boldsymbol{w}^T\boldsymbol{x}^i}} (\boldsymbol{x}^i)^T \boldsymbol{v_2} \nonumber \\
  &\geq \Lambda ||v_1||^2 + 2\mu ||v_2||^2 - 2 ||v_1||_1 ||v_2||_2 \nonumber \\
  &\geq \Lambda ||v_1||^2 + 2\mu ||v_2||^2 - 2 \sqrt{m} ||v_1||_2 ||v_2||_2
\end{align}
Since, $||v_1||^2 + ||v_2||^2 = 1$, let $||v_1||_2 = \cos \theta$ and $||v_2||_2 = \sin \theta$. We need to find a condition such that $\boldsymbol{v}^T \nabla^2 g \boldsymbol{v} > 0$ for every $\theta$. Therefore,
\begin{align}
    &\Lambda ||v_1||^2 + 2\mu ||v_2||^2 - 2 \sqrt{m} ||v_1||_2 ||v_2||_2 \nonumber \\
    &= \Lambda \cos^2 \theta + 2\mu \sin^2 \theta - \sqrt{m} \sin 2 \theta \nonumber \\
    &= \frac{\Lambda + 2\mu}{2} + \frac{\Lambda - 2\mu}{2} \cos 2 \theta - \sqrt{m} \sin 2 \theta.
\end{align}
Thus, for the above equation to be positive
\begin{align}
    &\frac{\Lambda + 2\mu}{2} > \sqrt{m + \frac{\Lambda-2\mu}{2}} \nonumber \\
    \implies &2\Lambda \mu > m.
\end{align}
Thus, $2\Lambda \mu > m$, is a sufficient condition for the loss function to be strictly convex.

Next, we proceed to prove the second part. Let the function be $\mu$-strictly convex.  Denote $(\boldsymbol{w},\boldsymbol{a})$ by $\boldsymbol{v}$, and let $\inf_{v} f(v) = L$. Then, there exists $v^{*} \in \mathbb{R}^{m+n}$, such that $\forall \ \delta > 0$
\begin{equation*}
    f(v^{*}) - L < \delta.
\end{equation*}
Thus, we have
\begin{align}\nonumber
    f(v) - L &= f(v) - f(v^{*}) + f(v^{*}) - L \nonumber 
    \end{align}
\begin{align}
    &\leq \delta + \langle \bigtriangledown f(v), v - v^{*}\rangle - \frac{\mu}{2} \|v^{*}-v\|^2 \cr
    &\leq \delta + \frac{1}{2\mu} \|\bigtriangledown f(v)\|^2.
    \label{PL}
\end{align}
Next, we prove that $f$ is $L$-smooth. Let $\bigtriangledown_{\boldsymbol{w}} f(v)$ and $\bigtriangledown_{\boldsymbol{a}} f(v)$  denote the gradient vectors of $f$ with respect to $\boldsymbol{w}$ and $\boldsymbol{a}$, respectively. Then
\begin{align}\nonumber
    & \|\bigtriangledown f(v^1) - \bigtriangledown f(v^2) \| ^2 \cr
    &= \|\bigtriangledown_{\boldsymbol{w}} f(v_1) - \bigtriangledown_{\boldsymbol{w}} f(v_2) \| ^2 + \| ( \log(1+e^{-\boldsymbol{w^1}^{T}\boldsymbol{x}y}) \nonumber \\
    &+ 2\mu \|\boldsymbol{a}^1\| -   (\log(1+e^{-\boldsymbol{w^2}^{T}\boldsymbol{x}y}) - 2\mu \|\boldsymbol{a}^2\|\|^2.
\end{align}
Now, the logistic loss is $L_1$-smooth for some $L_1 > 0$. Further, $||\boldsymbol{a}||$ is also $L_2$-smooth for some $L_2 > 0.$ Thus, there exists an $K$ such that $f$ is $K$-smooth. 

Consider that the step size for gradient descent is chosen such that, $\gamma K \leq 1$. Therefore, by descent lemma, we have
\begin{align}\nonumber
    f(v^{t+1}) &\leq f(v^t) + \langle\bigtriangledown f(v^t), v^{t+1} - v^t\rangle + \frac{K}{2} \|v^{t+1}-v^t\|^2 \cr
    &\leq f(v^t) - \gamma ||\bigtriangledown f(v^t)||^2 + \frac{K \gamma^2}{2} \|\bigtriangledown f(v^t)\|^2 \cr
    &\leq f(v^t) - \frac{\gamma}{2} \| \bigtriangledown f(v^t)\|^2.
\end{align}
The second inequality uses the condition for projection, and that projection is non-expansive. Finally, the update rule is substituted to obtain terms with $\| \bigtriangledown f(v^t)\|^2$. By using Eq (\ref{PL}) and applying recursion, we get for all $\delta > 0.$ 
\begin{equation}\nonumber
    f(v^t) - L \leq (1-\gamma \mu)^t (f(v^0)-L) + \delta.
\end{equation}

\end{proof}

\begin{figure}[H]    \centering
    \label{fig:eta_effect}
    \subfloat[\centering Algorithm performance with respect to $\eta$]{{\includegraphics[width=3cm]{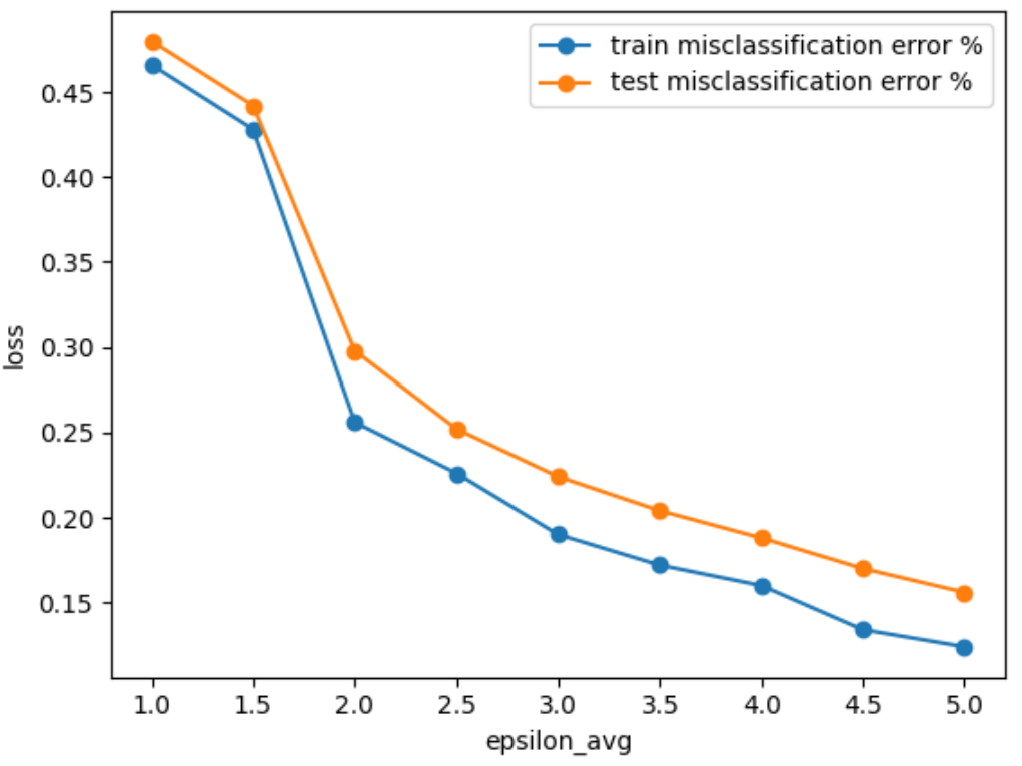} }}
    \qquad
    \label{fig:mu_effect}
    \subfloat[\centering Algorithm performance with respect to $\mu$]{{\includegraphics[width=3cm]{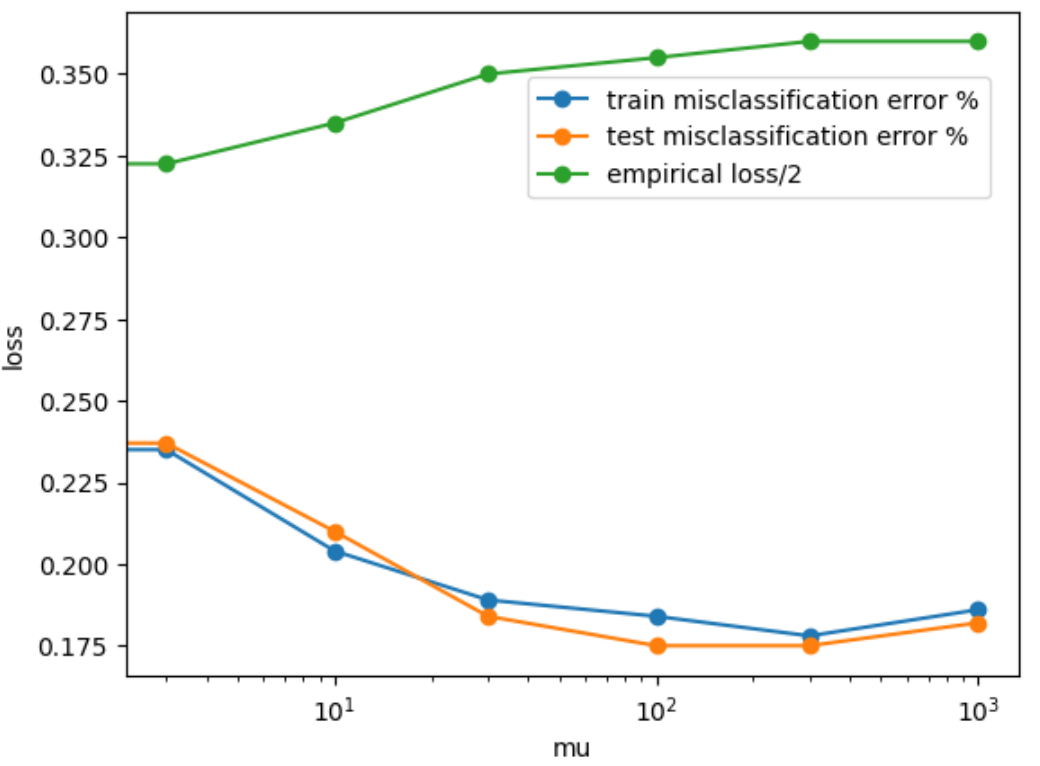} }}
    \caption{Hyperparameter Analysis}
    \label{fig: hyperparameter analysis}
\end{figure}
\section{Additional Experiments}
We perform additional experiments on generated synthetic data. For synthetic data, we consider the classification boundary to be a linear separator passing through the origin. Input data $\boldsymbol{x}^i$ is generated by sampling from i.i.d. zero-mean Gaussian distribution with bounded variance. Corresponding outputs, $y^i$, are generated using the linear separator. Furthermore, $\boldsymbol{c}$ is drawn from $\mathbb{U}[p,q]$, where $p,q \in \mathbb{R}$.

\subsection*{Intuition behind hyperparameter $\mu$}
We solve the logistic regression problem while ensuring heterogeneous differential privacy, i.e., Eq. (\ref{derived-loss-1}). The misclassification error is compared for different values of $\mu$. Results are plotted in Figure \ref{fig: hyperparameter analysis}. Train/test misclassification error is the percentage of misclassified samples while empirical loss is given by $\sum_{i=1}^{m} a_{i} \log(1+e^{-y^i \cdot \boldsymbol{w}^T \boldsymbol{x}^i})$. We see that as $\mu$ increases, there is a reduction in both train and test misclassification errors at first, and then it increases slightly. Moreover, we observe that for a smaller $\mu$, the empirical loss is small even if samples are misclassified. Therefore, optimizing over the empirical loss alone might not result in a good logistic regression model. Hence, it is also necessary to consider generalization terms in the optimization.



\subsection*{Performance with respect to parameter $\eta$}
 We fix $a_i = \frac{1}{m}$ and vary $\eta$. We write $\eta = m\cdot \epsilon_{avg}$ and vary $\epsilon_{avg}$. Therefore, we solve logistic regression while adding noise to ensure differential privacy.
\newpage

\begin{IEEEbiography}[{\includegraphics[width=1in,height=1.25in,clip,keepaspectratio]{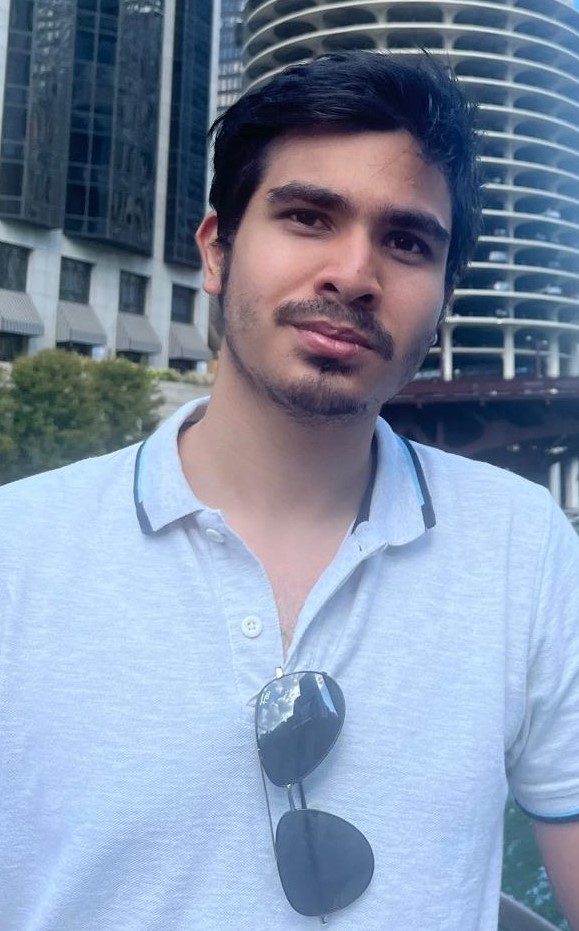}}]{Ameya Anjarlekar} received his B.Tech in Electrical Engineering from the Indian Institute of Technology, Bombay. He is currently pursuing his Ph.D. degree at the University of Illinois at Urbana Champaign in Electrical and Computer Engineering. He is a Research Assistant at the Coordinated
Science Laboratory, University of Illinois at Urbana-Champaign. His research interests include Game Theory, Privacy-Preserving Machine Learning and Reinforcement Learning for Partially Observed Systems.
\end{IEEEbiography}
\begin{IEEEbiography}[{\includegraphics[width=1in,height=1.25in,clip,keepaspectratio]{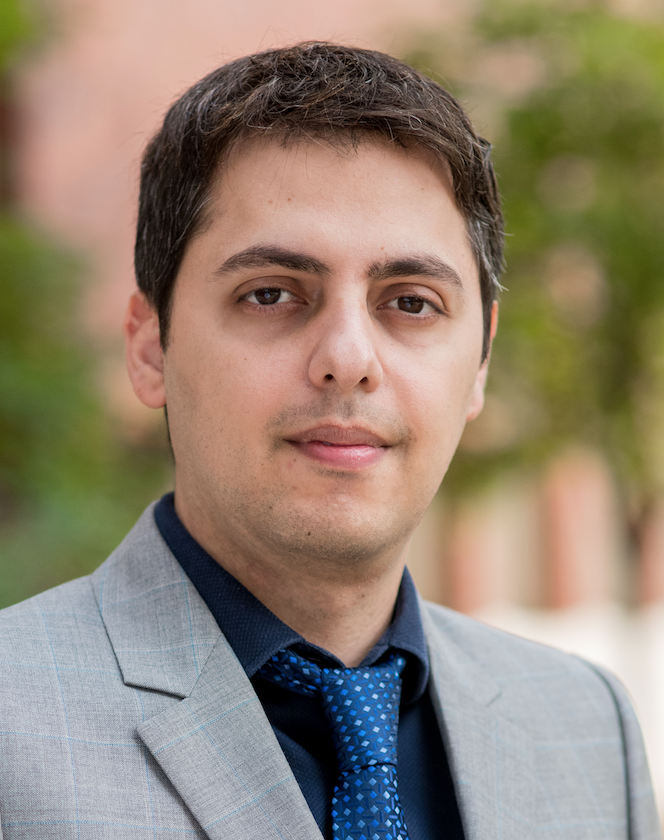}}]{Seyed Rasoul Etesami} is an Associate Professor in the Department of Industrial and Systems Engineering (ISE) at the University of Illinois Urbana-Champaign. He is also affiliated with the Department of Electrical and Computer Engineering (ECE) and the Coordinated Science Laboratory (CSL) at the University of Illinois Urbana-Champaign, where he is a member of the Decision and Control group. From 2016 to 2017, he was a Postdoctoral Research Fellow in the Department of Electrical and Computer Engineering at Princeton University and WINLAB. He received his Ph.D. in Electrical and Computer Engineering from University of Illinois Urbana-Champaign, during which he spent a summer as a Research Intern at Alcatel-Lucent Bell Labs.
\end{IEEEbiography}
\begin{IEEEbiography}[{\includegraphics[width=1in,height=1.25in,clip,keepaspectratio]{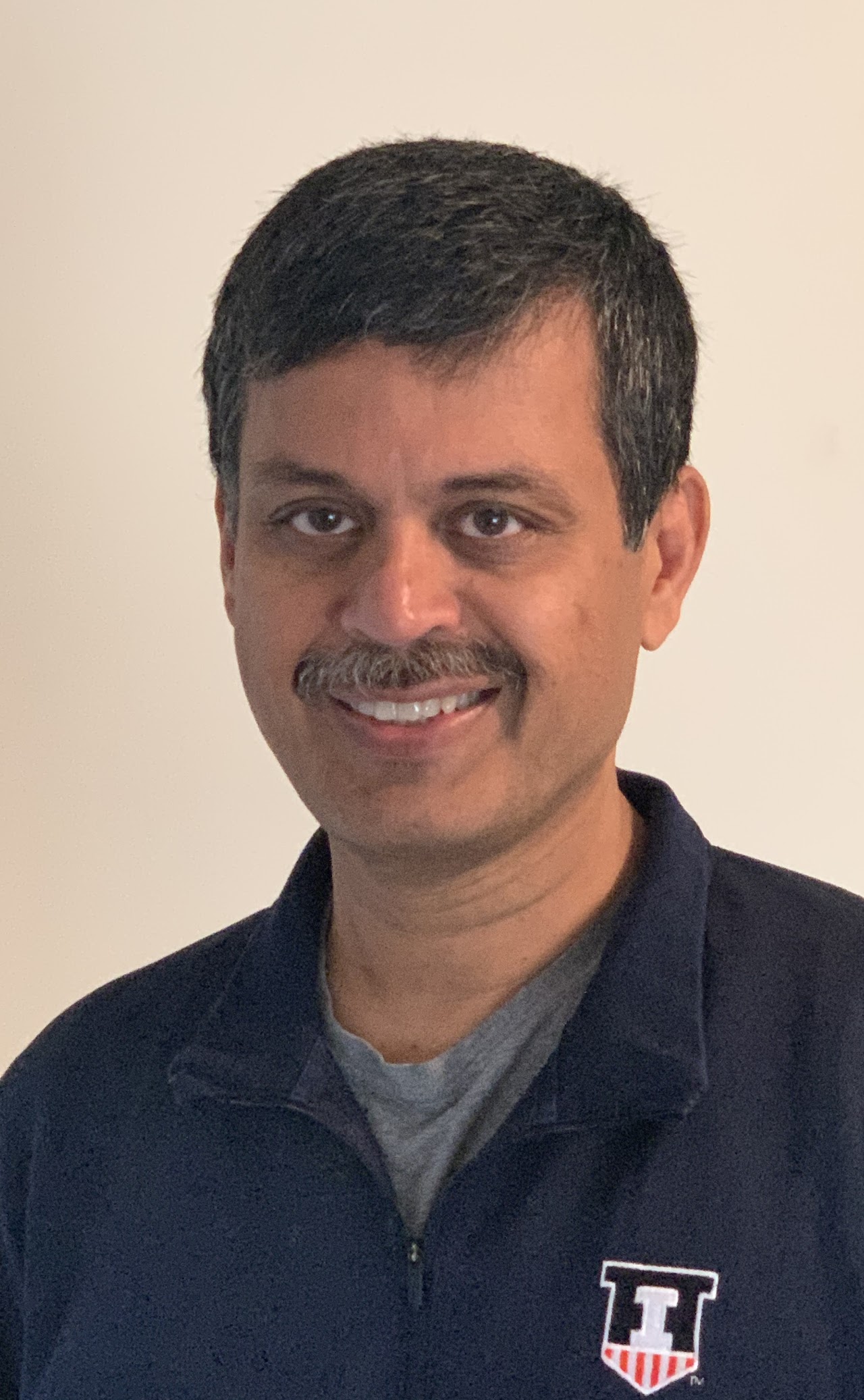}}]{R Srikant} (Fellow, IEEE)  received the B.Tech.
degree from the Indian Institute of Technology,
Madras, Chennai, India, in 1985, the M.S. and
Ph.D. degrees from the University of Illinois at
Urbana-Champaign, Champaign, IL, USA, in
1988 and 1991, respectively, all in electrical engineering.
He was a Member of Technical Staff with
AT\&T Bell Laboratories from 1991 to 1995. He is
currently with the University of Illinois at Urbana-Champaign, where he is the co-Director of the
C3.ai Digital Transformation Institute, a Grainger Distinguished Chair
in Engineering, and a Professor with the Department of Electrical and
Computer Engineering and the Coordinated Science Lab. His research
interests include applied probability, machine learning, stochastic control
and communication networks.
Dr. Srikant is currently the recipient of the 2015 INFOCOM Achievement Award, the 2019 IEEE Koji Kobayashi Computers and Communications Award, and the 2021 ACM SIGMETRICS Achievement Award.
He is also the recipient of several Best Paper awards including the 2015
INFOCOM Best Paper Award, the 2017 Applied Probability Society Best
Publication Award, and the 2017 WiOpt Best Paper award.
\end{IEEEbiography}

\end{document}